\def\year{2021}\relax
\documentclass[letterpaper]{article} 
\usepackage{aaai21}  

\onecolumn
\usepackage{times}  
\usepackage{helvet} 
\usepackage{courier}  
\usepackage[hyphens]{url}  
\usepackage{hyperref}
\usepackage{graphicx} 
\usepackage{natbib}  
\usepackage{caption} 
\usepackage{graphics} 
\usepackage{graphicx}\graphicspath{ {./figures/} }
\usepackage{amsmath} 
\usepackage{amssymb}  
\usepackage{bm}
\usepackage{multicol}
\usepackage{booktabs}
\usepackage[linesnumbered,ruled]{algorithm2e}
\usepackage{subcaption}
\usepackage{mathtools}

\DeclareMathSizes{6.5}{6.5}{6.5}{6.5}

\usepackage[para]{footmisc}
\usepackage{cite}
\usepackage{istgame}
\usepackage{subcaption}
\usepackage{algorithm2e}
\usepackage{amsmath}
\usepackage{multirow}
\usepackage{array}
\usepackage{textcomp}
\usepackage{siunitx}
\usepackage{adjustbox}
\usepackage{bm}
\usepackage{rotating}
\usepackage{placeins}
\usepackage{textcomp}
\usepackage{atbegshi}
\usetikzlibrary{decorations.pathmorphing}
\newcolumntype{P}[1]{>{\centering\arraybackslash}p{#1}}
\DeclareMathOperator*{\argminB}{argmin}
\DeclareMathOperator*{\argmaxB}{argmax}
\DeclareMathOperator{\erf}{erf}
\title{Solution Concepts in Hierarchical Games Under Bounded Rationality With Applications to Autonomous Driving}

\author{
    Atrisha Sarkar \textsuperscript{\rm 1},
    Krzysztof Czarnecki \textsuperscript{\rm 2}
    \\
}
\affiliations{
\textsuperscript{\rm 1}David R Cheriton School of Computer Science
\textsuperscript{\rm 2}Department of Electrical and Computer Engineering\\
University of Waterloo, Ontario, Canada\\

    atrisha.sarkar,k2czarne@uwaterloo.ca

}

\AtBeginDocument{
  \AtBeginShipoutNext{
    \AtBeginShipoutUpperLeft{
      \put(\dimexpr\paperwidth-20.5cm\relax,-1.7cm){
        \parbox[b]{20cm}{
          Extended version of ``Sarkar, A., \& Czarnecki, K. (2021). Solution Concepts in Hierarchical Games Under Bounded Rationality With Applications to Autonomous Driving. In Proceedings of the AAAI Conference on Artificial Intelligence (Vol. 35, Issue 6, pp. 5698–5708). Association for the Advancement of Artificial Intelligence (AAAI)"
        }
      }
    }
  }
}

\begin{document}

\maketitle

\begin{abstract}
With autonomous vehicles (AV) set to integrate further into regular human traffic, there is an increasing consensus on treating AV motion planning as a multi-agent problem. However, the traditional game-theoretic assumption of complete rationality is too strong for human driving, and there is a need for understanding human driving as a \emph{bounded rational} activity through a behavioural game-theoretic lens. To that end, we adapt four metamodels of bounded rational behaviour: three based on Quantal level-k and one based on Nash equilibrium with quantal errors. We formalize the different solution concepts that can be applied in the context of hierarchical games, a framework used in multi-agent motion planning, for the purpose of creating game theoretic models of driving behaviour. Furthermore, based on a contributed dataset of human driving at a busy urban intersection with a total of approximately 4k agents and 44k decision points, we evaluate the behaviour models on the basis of model fit to naturalistic data, as well as their predictive capacity. Our results suggest that among the behaviour models evaluated, at the level of maneuvers, modeling driving behaviour as an adaptation of the Quantal level-k model with level-0 behaviour modelled as pure rule-following provides the best fit to naturalistic driving behaviour. At the level of trajectories, bounds sampling of actions and a maxmax non-strategic models is the most accurate within the set of models in comparison. We also find a significant impact of situational factors on the performance of behaviour models.   

\end{abstract}

\section{Introduction}
Motion planners are a critical component of autonomous vehicle (AV) architecture, and the decisions made by the algorithms impact the safety of road users, such as pedestrians, cyclists, and other human-driven vehicles. Traditional approaches to motion planning have typically treated the problem as a single-agent problem; in this perspective, a vehicle interacts with the environment (in simulation or on-field setting), possibly with the help of recorded human-driven trajectories, and plans its actions by optimizing over its objectives while taking into account the dynamic obstacles in the vicinity \cite{schwarting2018planning, ilievski2019design}. However, in reality human driving is a complex system with a symbiotic relation among agents, where actions of a vehicle influence the future actions of other road users and vice versa. 
More recently, there has been a focus towards treating motion planning of AVs as a multi-agent problem with game-theoretic solutions to AV decision-making \cite{fisac2019hierarchical,sadigh2016planning,camara2018empirical,li2018game}. Such approaches can account for heterogeneous objectives in a group of vehicles in a traffic scene and identify equilibrium solutions that guide the actions of the AV. Given that the movement dynamics of a vehicle are in a continuous domain, it is intuitive to model the dynamics as a differential game, an approach adopted by multiple models in the literature \cite{fridovich2019efficient,sadigh2016planning,wang2019game}. However, the applicability of such games as a general purpose planner is limited by the trade-off between the computational burden and expressivity; cases where efficient solutions exist in a multi-agent setting restrict the behaviour of the agents to only linear dynamics \cite{fridovich2019efficient}. As an alternative, \cite{fisac2019hierarchical} introduced the concept of a hierarchical game for AV planning where the game is decomposed into two levels; a long-horizon strategic game that can model richer agent behaviour and a short-horizon tactical game with simplified information structure. Although hierarchical games are well suited and show promising results for planning in AV, for the models to be applicable in real world situations, we need to understand how well the stationary concepts in the game match naturalistic human driving behaviour. It is well known that in many realistic settings, the theoretical fixed point of Nash equilibrium is a poor predictor of human behaviour \cite{goeree2001ten}; therefore, it is necessary to investigate if the same is true for human driving behaviour too. In absence of that information, we do not know \emph{what} to optimize for.\par
Behavioural game theory provides a framework to analyze decision-making in a naturalistic setting, and such behaviour models often have higher predictive power than Nash equilibria \citep{camerer2011behavioral}. A key element in behavioural game theory is \emph{bounded rationality}, where the conventional game-theoretic notion of agents as fully rational is relaxed to allow for sub-optimal behaviour. Such behaviour may arise from limitations in cognitive reasoning, or error-prone actions \cite{samuelson1995bounded}. Driving is a cognitively demanding job that requires situational awareness and sophisticated visuomotor co-ordination, added on to individual habits, biases, and preferences; and it is not hard to imagine that driving at its core is a bounded rational activity. Consequently, it becomes essential for AV game theoretic planners to be able to characterize the bounded rational behaviour in human driving; for example, if humans are prone to making error in judgement when the signal is about to turn red from amber at a busy intersection, then the AV planner should take that into account since the safety of the AV decision is conditioned on the error made by the human driver. Therefore, developing a game-theoretic planner for an AV is a multi-step process, broadly involving a) selection of the right behaviour model and equilibrium concepts for other road agents, b) estimation of the parameters of the model, and c) generation of a safe maneuver and trajectory after accounting for the model and its parameters. In this paper, we primarily focus on the first two aspects.\par
Wright and Leyton-Brown developed a general framework of analysing and estimating parameters of popular behavioural game theory models based on observations of game play. They focus on two models of behaviour, i.e. Quantal Level-k (QLk) and Poisson-Cognitive Hierarchy (P-CH) \cite{wright2012behavioral}, which model iterated reasoning where agents have a limited capacity to maintain higher order belief about other agents. Although QLk and P-CH do not capture all types of bounded rationality that one can think of in the case of human driving, such as the ones that arise from sampling the actions of other agents, the framework developed in \cite{wright2012behavioral} nevertheless can be applied to a wider set of behaviour models including the ones we develop in this paper.\par
Developing a game-theoretic planner for an AV is a multi-step process, broadly involving a) selection of the right behaviour model and equilibrium concepts for other road agents, b) estimation of the parameters of the model, and c) generation of a safe maneuver and trajectory after accounting for the model and its parameters. In this  paper, we focus on the first two aspects. We also restrict the focus in this  paper to the single-shot moving horizon based setting, which is the planning process where agents play a fixed time horizon game at a constant planning frequency, starts execution of their action and replans again in the next time step. The contributions of this  paper are as follows. 
\begin{itemize}
    \item Formalisation of the concept of a hierarchical game along with the various solution concepts from behavioural game theory that can be applied to the solve such games.
    \item Development and evaluation of thirty behaviour models demonstrating different methods of game construction and solution concept choices for modelling traffic interactions.
\end{itemize}
\par

\section{Related work}
One of the first works to include game theory as a methodology to model human driving behaviour is by Kita \citep{kita1999merging}, where a lane change scenario is modelled as a two player game. The solution concept used in that work is a mixed strategy Nash equilibria of merge/give way behaviour. MLE (Maximum Likelihood Estimation) is used to estimate utility parameters based on data recorded by a video camera on a Japanese highway. Since \citep{kita1999merging}, many game theoretic models have focused on lane change behaviour, and a recent review provides good coverage of this literature \citep{ji2020review}. Since the interest of this paper is autonomous vehicles, we review the relevant literature in that domain of application in more detail. We categorize the relevant literature along key dimensions as shown in Table \ref{choneshot:relev_lit}.
\begin{itemize}
    \item \textit{Structure}: This refers to the structure of the game as modelled by the available actions and the planning time horizon. Most of the works, including ours in this  paper, are of one shot games with moving horizon \citep{yu2018human, zimmermann2018carrot, estivill2019game, li2016hierarchical, liniger2019noncooperative, li2020game, Geiger_Straehle_2021, geary2020resolving, garzon2020game}. In this type of construction, agents play a normal form game (therefore the name one shot) constructed with respect to a fixed horizon. Based on the solution of the game, the agent starts to proceed with their action, and replan again by constructing a new game at fixed time interval, which is the planning frequency. This method of planning with a moving or receding horizon, in which only a part of the planned action is executed before re-planning again, is commonly used in AV motion planning techniques \citep{claussmann2019review}. An alternate way of solving the game is replacing the normal form game with a dynamic game, thereby supporting a richer strategy and action space. In this construction, the agent repeatedly plays a dynamic game, executes the action, and replans again at the terminal node. In a single-agent setting, this method of planning is similar to a Model Predictive Control (MPC) based planning \citep{camacho2013model}. Since the construction of the dynamic game may involve an exponentially large state space, Monte Carlo-based sampling methods are often used in the construction and evaluation of the game tree \citep{tian2021anytime, sun2020game}. One can refer Sarkar et al. \citep{sarkar2022generalized} for planning in AV using solution concepts for dynamic games.
    \item \textit{Solution concept} : This categorisation is based on the solution concept used in solving the game. Stackelberg and Nash equilibrium are more commonly used solution concepts in this regard, and more recently, level-k \citep{tian2018adaptive} type methods have also been in use in order to support boundedly rational agents. Stackelberg solution lends well to the problem of traffic interaction, since many situations can be modelled as one agent being the leader and the other as the follower. For example, the driver holding the right of way can be modelled as a leader. However, since this assignment is part of common knowledge and has to be agreed upon by both agents, the use of Stackelberg may be too restrictive due to this assumption. Geary et al. \citep{geary2020resolving} show that the breakdown of this assumption can lead to dangerous situations (such as collisions and stopping on highways) and show that changing the utility structure to model aspects such as altruism can be one way to avoid such situations. The solution concept used in this  paper is based on a hierarchical decomposition of the game and is different for different levels of the game. Such a hierarchical decomposition with respect to the solution concept is seen in the work by Fisac et al. \citep{fisac2019hierarchical}, and more recently, the use of local Nash equilibrium in \citep{Geiger_Straehle_2021} can also be interpreted as a form of hierarchical decomposition. 
    \item \textit{Empirical} : Although there have been several sophisticated models developed within a game theoretic setting, since the problem is one of modelling human behaviour, it is vital to judge the effectiveness of the models in real world setting. Therefore, we cover the literature on the basis of whether the proposed methods include empirical evaluation. Some works include evaluation only in simulation where the efficacy is demonstrated by performance in different scenarios in simulation. Other methods include experiments with human subjects in a driving simulator where either data is collected based on how humans drive in the simulator or models are evaluated against a human taking the role of one of the players. However, there is a gap in the literature on the evaluation of models based on naturalistic driving data in a real world setting, which the work in this paper aims to address. Contemporary works that address the proposed game-theoretic models based on naturalistic data include \citep{sun2020game} and \citep{Geiger_Straehle_2021}, both of which were published in the similar time frame as this  paper.
    \item \textit{Bounded rationality}: This categorization is based on whether the models have support for boundedly rational agents. Most game theoretic models are built upon the assumption that agents are completely rational, either through their ability to calculate a best response action often in conjunction with their ability to reason over the possible actions of other agents. This can be taken as a reasonable assumption when models are evaluated based on human in simulator studies, since it is possible to give all the relevant information to the participant, which includes the utilities, the actions, game structure, etc. However, in a real-world setting all of these elements are outside of the control of the game modeller, and therefore having support for boundedly rational agents (bounded from the perspective of the game modeller) becomes essential.
    \item \textit{Scenario} : This column refers to the specific traffic scenario based on which the models are developed or evaluated. Typical examples include the lane change scenario (which has been the most common scenario studied), along with the scenarios of intersection and roundabout in recent years.
    \item \textit{Utilties} : Driving is often a multi-objective activity that includes balancing multiple objectives such as safety, progress, comfort, and most of the literature shown in Table \ref{choneshot:relev_lit} is reflective of that. Typical dimensions along which the proposed methods model driver behaviour include safety, progress, comfort, adherence to lane and speed limits, along with behavioural attributes such as empathy and altruism in some cases. Since solution concepts needed to solve the games involve aggregation of the multivalued utility into a single real number, a linear weighting of the objectives is the most common method of aggregation. In some cases, the weights are estimated \emph{a-priori} through a separate process based on techniques such as Inverse Reinforcement Learning \citep{tian2021anytime}, and in other cases fixed weights are also used. Alternatively, utilities can also be modelled purely through demonstrations as in \citep{sadigh2016planning}, where any action that is more similar to the demonstrated action fetches higher utility and, therefore, the canonical dimensions of safety, progress, etc. are not modelled explicitly. 
 \end{itemize}
 In relation to the literature presented above, the methods developed in this  paper fall under one-shot moving horizon in terms of the game structure, solution concepts include Nash equilibrium, Level-k, along with elementary non-strategic decision models, and intersection as the scenario of study. The methods also focus on modelling boundedly rational behaviour, and we evaluate the models based on naturalistic driving data.
\begin{sidewaystable*}
\centering
\normalsize
\resizebox{\textheight}{!}{%
\begin{tabular}{p{2cm}cp{2cm}p{2.5cm}cccc}
\toprule
& Year & Structure           & Solution concept & Evaluation                 & Bounded rationality & Scenario    & Utilities              \\ \midrule
Yu et al.. \citep{yu2018human} & 2018 & One shot moving horizon & Stackelberg      & In simulator, simulation. & No                  & Lane change & Distance gap, Time gap \\ 
Zimmerman et al.. \citep{zimmermann2018carrot}& 2018 & One shot moving horizon & Nash      & In simulator & No                  & Lane change & Cooperative utility, time pressure, safety. \\ 
Castro \citep{estivill2019game}& 2019 & One shot moving horizon & Mixed strategy Nash      & Simulation only & No                  & Trolley problem & Safety. \\ 
Li et al.. \citep{li2016hierarchical}& 2019 & One shot moving horizon & Stackelberg  & Simulation only & No                  & Lane change & Safety, progress, effort. \\ 
Li et al.. \citep{li2019game}& 2019 (2016) & Dynamic game & Level-k      & Simulation only & Yes                  & Lane change & Progress, safety, effort. \\ 
Sadigh et al.. \citep{sadigh2016planning}& 2016 & MPC & Level-k & In simulator, simulation  & Yes                  & Lane change & Learned through IRL. \\
Fisac et al.. \citep{fisac2019hierarchical}& 2019 & Dynamic & Stackelberg, maxmax & Simulation only  & Yes                  & Lane change & Safety, progress. \\
Liniger and Lygeros \citep{liniger2019noncooperative}& 2019 (2017) & One shot moving horizon & Nash, Stackelberg      & Simulation only & No                  & Racing & Progress, safety, \emph{blocking}. \\
Sun et al.. \citep{sun2018courteous}& 2018 & MPC & Nash& Simulation, utility estimation on NGSIM & No                  & Lane change, intersection & IRL based. \\
Li et al.. \citep{li2020game}& 2020 & One shot moving horizon, Dynamic. & Stackelberg, Level-k& Partially naturalistic (selected behaviour replication) & Yes                  & Uncontrolled intersection & Safety, progress. \\
Geiger and Straehle \citep{geiger2020learning}& 2020 & One shot moving horizon & local Nash equilibrium & Naturalistic & No                  & Lane change & Progress, speed limit adherence. \\
Geary et al.. \citep{geary2020resolving}& 2021 & One shot moving horizon & Stackelberg & In simulator, simulation  & No                  & Lane change & Progress, speed limit adherence. Altruism. \\
Tian et al.. \citep{tian2021anytime}& 2021 & Dynamic game (Monte Carlo) & Level-k & In simulator, simulation  & Yes                  & Lane change & Safety, progress, comfort. \\
Sun et al.. \citep{sun2020game}& 2020 & Dynamic game (Monte Carlo) & Nash, Stackelberg, Pareto & Naturalistic  & No                  & Roundabout & Safety, speed limit adherence. \\
Garz{\'o}n and Spalanzani \citep{garzon2020game}& 2020 & One shot moving horizon & Level-k & Simulation only  & No                  & Merge & Goal completion. \\
This  paper& 2020 & One shot moving horizon & Level-k (with variations), Nash equilibrium with Quantal Errors & Naturalistic   & Yes               & Intersection & Safety and progress \\
\bottomrule
\end{tabular}
}
\caption{Relevant literature on the application of game theoretic models for autonomous vehicles.}
\label{choneshot:relev_lit}
\end{sidewaystable*}

\section{Hierarchical Games}

Prior to the recent focus on autonomous driving, there has been a considerable body of research on modelling driving behaviour within the field of traffic psychology with a long history of treating driving behaviour as a hierarchical model \citep{keskinen2004driver,van1988hierarchical,lewis2012testing,michon1985critical}. One of the more influential models, the Michon hierarchy of driving tasks \citep{michon1985critical}, decomposes driving into three levels of control; a strategic plan such as a route and general speed choice of going from point A to B is decomposed into several tactical decisions of choosing the right manoeuvres, which is further decomposed into high-fidelity actions that control the steering and acceleration. 
A primary motivation of a hierarchical decomposition is that drivers have different motivations and risk judgements at each level of the hierarchy, and the functional decomposition into a hierarchical system allows modelling of risk and safety considerations separately at each level. A driver for example may be indifferent about the choices at a granular level of trajectories but care more about choosing the right manoeuvre, e.g., waiting for an oncoming vehicle. Motion planners in autonomous vehicles also follow a similar hierarchical pattern of decomposition; a high level \emph{route planner} plan is given to a \emph{behaviour planner}, which sets up the tactical manoeuvres for a lower level \emph{trajectory planner}, which in turn generates the trajectory profile for the vehicle \emph{controller} after respecting its nonholonomic constraints. In addition to the motivation mentioned above, treating the problem of planning as a hierarchical system is also driven by computational efficiency, as previously shown in \citep{fisac2019hierarchical}.

\subsection{Illustrative example}
\begin{figure*}[htb]
\centering
\includegraphics[width=.5\textwidth]{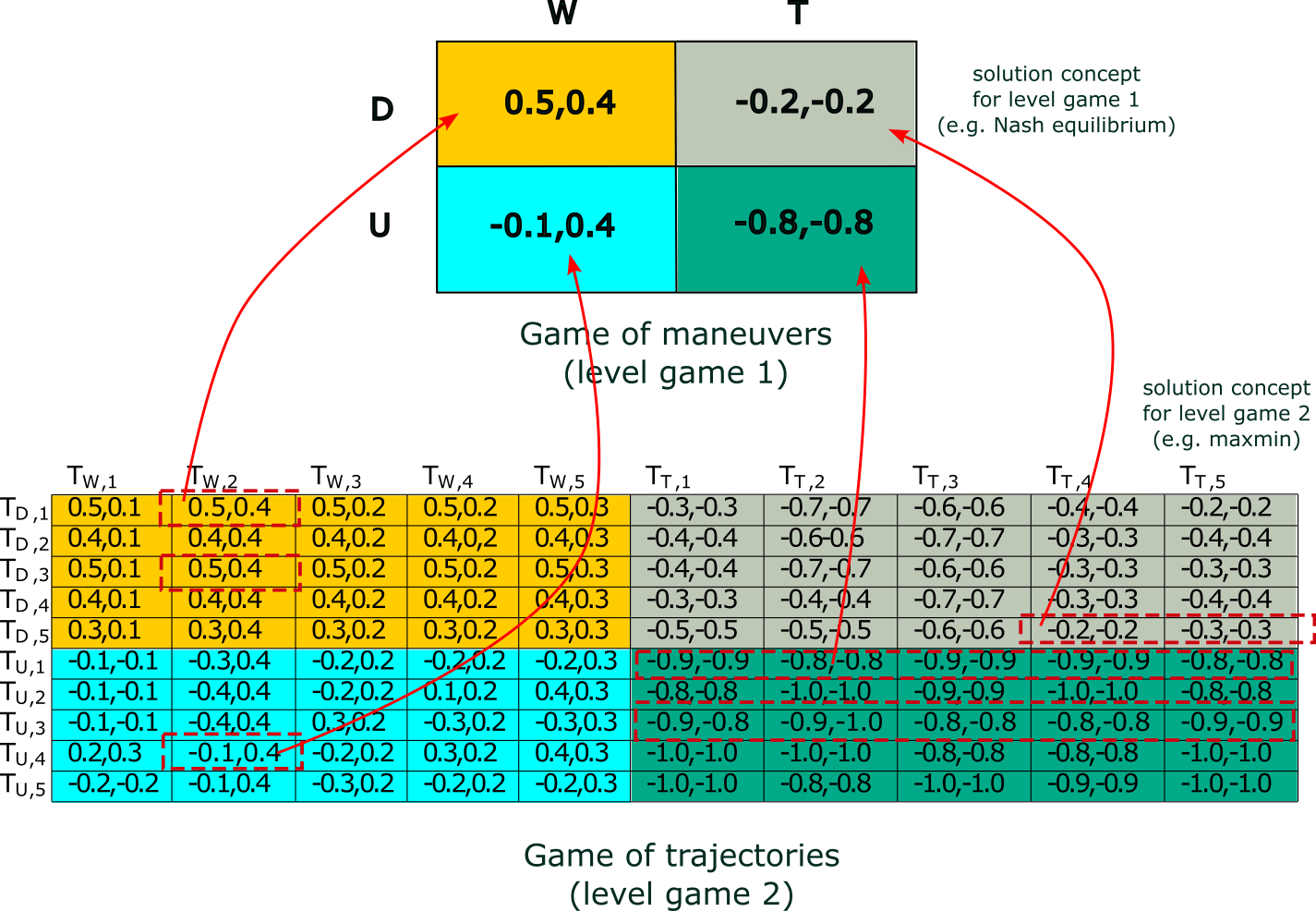}
  \caption{An example of a two level hierarchical game with action level game 1 being the game of manoeuvres and action level game 2 is the game of trajectories. Different solution concepts can be used at different levels to find a game solution.}
 \label{fig:hier_game_matrix}
\end{figure*}
First, we explain the construction of a hierarchical game through a simple illustrative example in this section, followed by a more formal construction in the next section. Fig. \ref{fig:hier_game_matrix} shows an example hierarchical game played between a vehicle turning right (column player) and a straight-through vehicle (row player) in an unprotected right turn at an intersection. This scenario is similar to the one shown in Fig. \ref{fig:scene} where the vehicle with id:14 is the right turning vehicle and the vehicle with id:26 is the straight-through vehicle. For the sake of simplicity of this example, we show only a 2 player game between 14 and 26 rather than an N player game involving all the relevant vehicles in the scene. Let us say the right turning vehicle has two high level manoeuvres available to them, \emph{wait} (W) or \emph{turn} (T), and similarly the straight through vehicle has two manoeuvres \emph{slow down} (D) or \emph{speed up} including the special case of maintaining their current speed (U). The high-level manoeuvre determines the velocity profiles, and then sets up the constraints expressed in terms of the target velocity ranges for a low-level trajectory planner (refer to the appendix for a detailed description of the trajectory generation process). For example, if $v_0$ is the vehicle speed during the initiation of the game, the slowing down manoeuvre may involve setting a constraint for the trajectory generation process with a target velocity in the range 
$[v_{T}^{\text{min}},v_{T}]$ where $v_{T} < v_{0}$ and $v_{T}^{\text{min}}$ is the minimum velocity reachable by the vehicle in $T$ seconds after taking into account the kinematic limits of the vehicle. Assuming that the vehicles use a method for sampling the possible trajectories (elaborated later in the paper), the trajectories are the main actions that the vehicles can execute in the game. Therefore, the set of trajectories for both vehicles forms the game matrix, as shown in the lower matrix of Fig. \ref{fig:hier_game_matrix}. Each coloured section of the matrix represents the trajectories corresponding to a specific combination of high-level manoeuvres. For now, let us assume that the utilities in the table are calculated after taking into account the various objectives such as safety, progress, etc. In the worst case scenario, finding a pure strategy Nash equilibrium of the game involves a quadratic time algorithm in terms of the size of the matrix that grows exponentially with the number of players \citep{ryan2010computing}. As an alternative, one can use a different solution concept that runs in linear time, e.g. maxmax (selection of the utility maximizing action), or does not involve pairwise comparison, e.g. maxmin (action that maximizes the utility of worst case scenario with respect to other agents' actions), for the subgame under each manoeuvre combination. Each shaded matrix in the game of trajectories can then be replaced by a representative solution (in this example we take a sample maxmin solution that maximizes the sum of utilities), thereby forming the game of manoeuvres as shown in the top matrix in the figure. The game of manoeuvres can subsequently be solved using a more computationally involved solution concept such as a pure strategy Nash equilibrium. The key idea behind hierarchical game construction is the use of heterogeneous solution concept at different levels based on the hierarchy of actions (manoeuvres and trajectories in our case). This is akin to the possible mental process involved in driving, where a driver deliberates more at the level of manoeuvres but less so at the level of individual trajectories within each manoeuvre combination. 
\subsection{Formalization}
\label{ch:one_shot:formalization}
\begin{figure}[h]
\centering
\begin{subfigure}[b]{0.2\textwidth}
\centering
\includegraphics[width=.95\columnwidth]{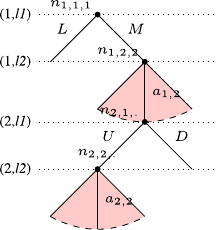}
\caption{}
\label{fig:stackelberg_tree}
\end{subfigure}
\begin{subfigure}[b]{0.24\textwidth}
\centering
\includegraphics[width=.95\columnwidth]{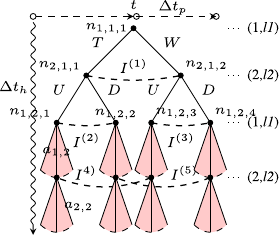}
\caption{}
\label{fig:our_tree}
\end{subfigure}
\caption{Illustration of two instances of hierarchical games. (a) As a Stackelberg game modelling a lane change maneuver and (b) simultaneous move game modelling intersection navigation. A hierarchical game is instantiated every $\Delta t_{p}$ seconds with action plan of $\Delta t_{h}$ seconds.}
\label{fig:static_hier}
\end{figure}
\begin{figure}[t]
\centering
\includegraphics[width=.6\columnwidth]{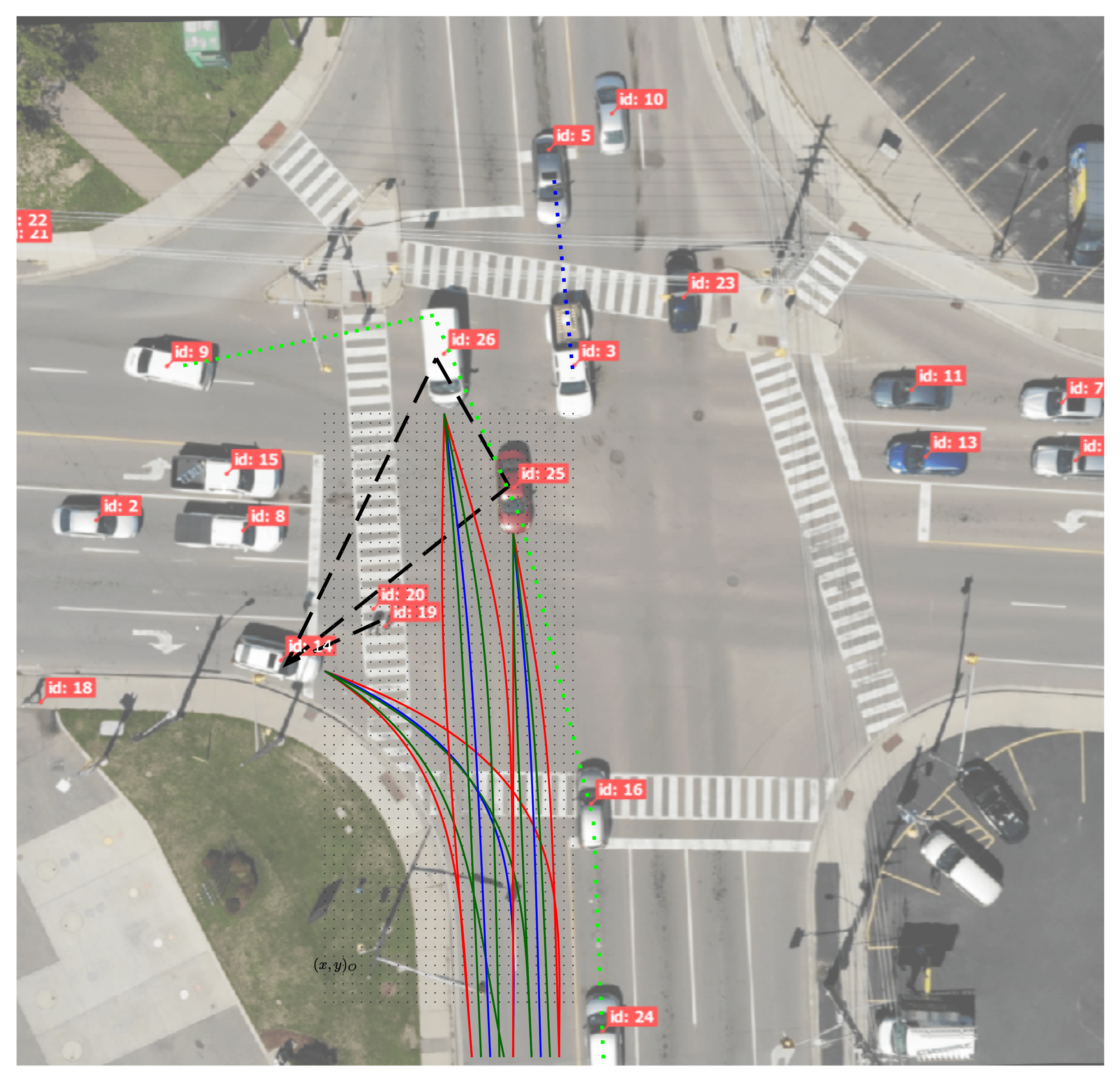}
  \caption{A snapshot of the intersection traffic scene. Representative trajectories based on the three sampling schemes over a $R^3$. The figure shows the path ($R^2$) projection of the trajectories and the dimension of time not represented in the figure.}
 \label{fig:scene}
\end{figure}

In this section, we formalise the construct of a hierarchical game for the general case of $N$ players and $\mathcal{K}$ levels of action hierarchy. Wherever possible, we use the term \emph{action level} to disambiguate between the levels of action hierarchy, and levels of cognitive hierarchy that appears later in the  paper as part of the behaviour models to solve the games. we also highlight the relation between solving a hierarchical game in the way that was illustrated in the example above and subgame solving through backward induction.
A hierarchical game is formulated by
\begin{itemize}
    \item Set of $N$ agents indexed by $i \in \{1,2,3,..N\}$.
    \item A set of $\mathcal{K}$ levels indexed by $\kappa \in \{1,2,3,..\mathcal{K}\}$.
    \item Set of actions $A_{i,\kappa}$ available to each agent $i$ at level $\kappa$.
    \item A strategy $s_{i}$ for agent $i$ is a $\mathcal{K}$-tuple $s_{i} = (a_{i,1},a_{i,2},..,a_{i,\mathcal{K}})$ where $a_{i,\kappa} \in A_{i,\kappa}$ and the strategy space of $s_{i}$ is $\displaystyle \prod_{\kappa \in \mathcal{K}} A_{i,\kappa}$.
    \item A set of states $X_{i}$ of agent $i$ in level 1, and an initial mapping function $f_{i,1}:X_{i} \rightarrow \mathcal{P}(A_{i,1})$ that maps the initial state of the agent to the available actions in level 1, where $\mathcal{P}(\cdot)$ is the power set.
    \item Set-valued functions $f_{i,\kappa} : {\displaystyle \prod_{j=1}^{\kappa-1} A_{i,j}} \rightarrow \mathcal{P}({A_{i,\kappa}})$ for each agent $i$ that maps a partial strategy ($a_{i,1},a_{i,2},..,a_{i,\kappa-1}$) to $\mathcal{P}(A_{i,\kappa})$ and gives the set of available actions to $i$ in level $\kappa > 1$ for the partial strategy till level $\kappa-1$.
    \item Set of $N$ pay-off (utility) functions $U=\{u_{i}(s_{i},s_{-i})\}$, where $-i$ refers to all agents other than $i$.
\end{itemize}
The hierarchical game imposes a total ordering in actions $A_i = \{A_{i,1},A_{i,2},..,A_{i,\mathcal{K}}\}$ of a given agent, and along with $f_{i,\kappa}$ induces a game tree, as shown in Fig. \ref{fig:our_tree}. The frequency at which a hierarchical game is instantiated ($\Delta t_{p}$) and the time horizon of each strategy ($\Delta t_{h}$) are exogenous to the model. Each node is labeled $n_{i,\kappa,j}$, where $i$ and $\kappa$ are the agent and level indices, and $j$ is the node identifier within level $\kappa$. This general formulation of a hierarchical game does not prescribe a fixed information structure, and allows the modeller to set an information structure that is appropriate to the environment and situation they want to model.  For example, Fisac et al. \citep{fisac2019hierarchical} models a lane change scenario where an AV merges into a lane occupied by a human driven vehicle. The game is modelled as a Stackelberg game with the AV being the leader and the human driven vehicle responding to the action of the AV. Fig. \ref{fig:stackelberg_tree} shows a 2-agent 2-level the game tree for such a scenario where each decision node is a singleton information set since for every decision node, the agent who owns the decision node has perfect information on where they are in the game. At node $n_{1,1,1}$, the AV (indexed as agent 1) has the choice of either staying in its current lane (L) or merging into the adjoining lane (M) $A_{1,1} = \{L,M\}$. Conditioned on this choice, the vehicle has to generate a trajectory $a_{1,2} \in f_{1,2}(a_{1,1})$ to execute the maneuver chosen in level 1. Whereas actions in $A_{1,1}$ are discrete choices, the agent can choose from a continuum of actions (shaded region in the figure) at node $n_{1,2,2}$. 
The human driven vehicle after having observed the actions of the AV, can respond by deciding to speed up (U) to dissuade the merging AV cut-in the front, or slow down (D) followed by a trajectory that corresponds to the choice. In situations where assignment of a leader and a follower is unclear or that assumption is too strong, the agents might not have perfect information on the state of the play. Interactions at an intersection for example, are such scenarios. Continuing from the illustrative example from the previous section, Fig. \ref{fig:our_tree} illustrates a 2-agent 2-level scenario as an example where an AV (indexed as 1) executes a free right turn on red at a signalized intersection (in a situation similar to id:14 in Fig. \ref{fig:scene}), while a human driven vehicle (id:26 and re-indexed as 2 in Fig. \ref{fig:our_tree}) approaches cross path from left to right. The AV can either decide to turn (T) or wait (W) for the cross path vehicle to pass, i.e., $f_{1,1}(X_{1})=A_{1,1}=\{T,W\}$. The human driven vehicle (id:26) can either slow down (D) or choose not to slow down (U), $f_{2,1}(X_{2})=A_{2,1}=\{D,U\}$. Since either agent does not have perfect information about what the other agent is about to do next, agent 2 does not know whether they are in node $n_{2,1,1}$ or $n_{2,1,2}$ (connected by the information set $I^{(1)}$). This imperfection of information is also reflected at the trajectory level (level 2 actions), where each agent can only distinguish between the nodes in level 2 that follow from their own chosen actions in level 1, but not from the ones that follow from the other agent's level 1 decision ($I^{(2)}$-$I^{(5)}$). \par 
It becomes apparent from this structure that the game has no proper subgame, and the game reduces to a simultaneous move game. It is well understood that a way to solve such games is by reduction to normal form. However, as we shall see, the hierarchical game has additional constraints that allow solving the game in Fig. \ref{fig:static_hier} also through backward induction. To designate the nodes where utilities accumulate at each level in the backward induction process, we label a set of nodes in each level $\kappa$ as \emph{level roots} $\mathcal{L}(\kappa)$ = $\{n_{i,\kappa,j}|parent(n_{i,\kappa,j}) \notin \mathcal{N}_{\kappa}\}$ where $\mathcal{N}_{\kappa}$ is the set of nodes in level $\kappa$. In other words, the set of level roots contain nodes in each level $\kappa$ whose parent is not in level $\kappa$. Therefore, $\mathcal{L}(1) = \{n_{1,1,1}\}$ and $\mathcal{L}(2) = \{n_{1,2,1},n_{1,2,2},n_{1,2,3},n_{1,2,4}\}$. Algorithm \ref{algo:back_ind} shows the standard backward induction process adapted to the hierarchical game.
\begin{algorithm}[!t]
\SetAlgoLined
\KwResult{$S_{1}^{*},V_{1}^{*} $}
 \For{$\kappa:= \mathcal{K}$;$\kappa=1$; $\kappa := \kappa - 1$ }{
 \For{$n \in  \mathcal{L}(\kappa)$}{
    $S^{*}_{\kappa,n},V^{*}_{\kappa,n} \leftarrow \text{solve}\ \mathcal{G}_{\kappa}({\displaystyle \prod_{i=1}^{N}} f_{i,\kappa}(\sigma_{i}(n))$,\\$\kappa = \mathcal{K} \mathord{?}  U;V_{\kappa+1,\mathcal{L}({\kappa+1})}^{*})$
   
 }
 }
\caption{Backward induction for a hierarchical game}
\label{algo:back_ind}
\end{algorithm}
The algorithm starts at the bottom most level ($\mathcal{K}$) and recursively moves up the tree by solving the action level games $\mathcal{G}_{\kappa}$ at every level. At each level, a simultaneous move \emph{action level game} $\mathcal{G}_{\kappa}$ is instantiated from each node in $\mathcal{L}(\kappa)$. These action level games are constructed by first extracting $\sigma_{i}(n)$, which gives the partial pure strategy for agent $i$ that lies on the branch from the root node of the game tree $\mathcal{L}(1)$ to node $n \in \mathcal{L}(\kappa)$. $f_{i,\kappa}$ gives the available actions for each agent $i$ in the current level $\kappa$, and these actions form the domain of available strategies in the action level game $\mathcal{G}_{\kappa}$. The utilities depend on the level of the game; for action level game $\mathcal{G}_{\kappa=\mathcal{K}}$ the utilities are same as the game utility $U$, whereas for action level games $\mathcal{G}_{\kappa < \mathcal{K}}$ are solved based on the game values $V^{*}_{\kappa+1,\mathcal{L}({\kappa+1})}$ from the game $\mathcal{G}_{\kappa+1}$ solved in the previous iteration. Note that the pseudocode shows only the case where a single solution and game value ($S^{*}_{\kappa,n},V^{*}_{\kappa,n}$) is propagated up the hierarchy. In the case of multiple solutions for the action level games, the strategies and values have to be tracked and repeated for each solution. The solutions and game value $S^{*}_{\kappa,n}, V^{*}_{\kappa,n}$ depend on the solution concept used for the individual action level game, and this is discussed in detail later under Solution concepts.\par
Due to the tree-like structure of the hierarchical game and the process of solving the game bottom up from the leaf nodes through backward induction, one can see that the process is very similar to solving for subgame perfect equilibria in multi-stage games with stages being replaced by levels in the hierarchy \citep{tadelis2013game}. However, it is not a subgame perfect equilibria, since the action level games are not subgames in the game tree. The connection between subgame perfect equilibria and the solution in the hierarchical game is that in the hierarchical game, the mapping functions $f_{i,\kappa}$ impose an action selection method that is similar to action selection based on the condition of \emph{sequential rationality} in subgame perfect equilibria. The mapping functions $f_{i,\kappa}$ eliminate strategies for all agents $i$ that are not direct successors of the partial strategies $\sigma_{i}(n) \cdot \sigma_{-i}(n)$, essentially breaking any information set within a level $\kappa$ that spans two separate levels in $\mathcal{L}(\kappa)$. To illustrate this more intuitively, take the example in Fig. \ref{fig:static_hier}(b) when agent 1 is in the information set $I^{(2)}$ connecting the two nodes $n_{1,2,1}$ and $n_{1,2,2}$. At this information set, agent 1 could solve the game under two nodes separately only under a guarantee that after solving the game starting from node $n_{1,2,1}$ and following that strategy, under no condition would they find themselves at a leaf node of the tree starting at $n_{1,2,2}$, and vice versa. The function $f_{i,\kappa}$ provides that guarantee and eliminates hypothetical strategies where, at level 1, a vehicle may think about slowing down, but at level 2 chooses a trajectory that speeds up; and the fact that this cannot happen is part of the common knowledge among the agents in the game mediated by $f_{i,\kappa}$.
\section{Game Structure} 
\label{sec:struct}
In this section, we describe the details of the game structure, including the number of agents, actions, strategies, and utilities.\par 
\subsection{Relevant agents and available actions} Since we are interested in investigating decision making in situations where there may be strategic reasoning involved, we extract situations where vehicles are executing a left turn across path or unprotected right turns. At each time step $\Delta t_{p}$=1s, we setup a hierarchical game with an action plan horizon of $\Delta t_{h}$=5s into the future from the perspective of each vehicle that turns left or right in the scenario. For example, in the snapshot of Fig. \ref{fig:scene}, the black dashed line shows the game from the perspective of the vehicle 14, which is turning right. The process of including the relevant vehicles in the game is as follows: we identify the conflict points on the map with respect to all the lanes in the intersection that cross each other. Since a game is initiated with respect to a ‘subject vehicle’, we first locate the conflict points corresponding to the lane that the subject vehicle is currently in, and include all agents in the scene that are on the lanes in conflict with the subject vehicle’s lane. We also include the leading vehicle of these conflicting vehicles. This set of relevant vehicles along with the subject vehicle form the set of agents in each game. Pedestrian actions are not modelled explicitly in the game tree; however, their influence is modelled in the utility structure of the game, which is described later in the section. \par
\begin{figure}[h]
\centering
  \begin{subfigure}[b]{0.45\columnwidth}
    \includegraphics[width=\linewidth]{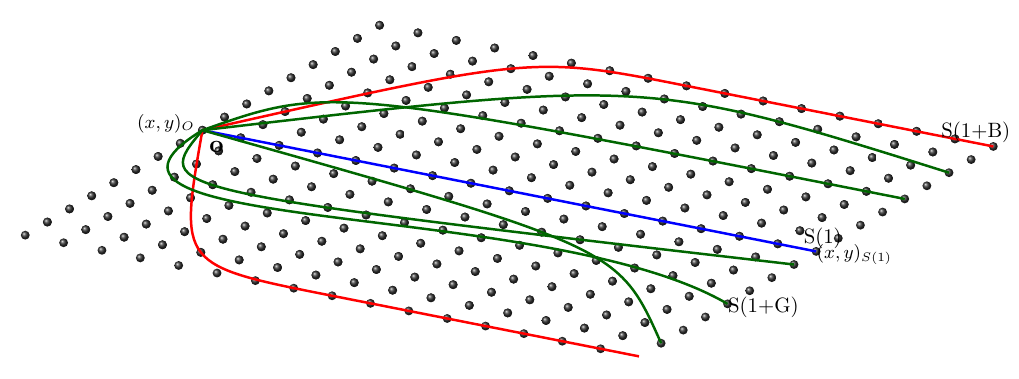}
    \caption{}
    \label{fig:lattice_path}
  \end{subfigure}
  \begin{subfigure}[b]{0.3\columnwidth}
    \includegraphics[width=\linewidth]{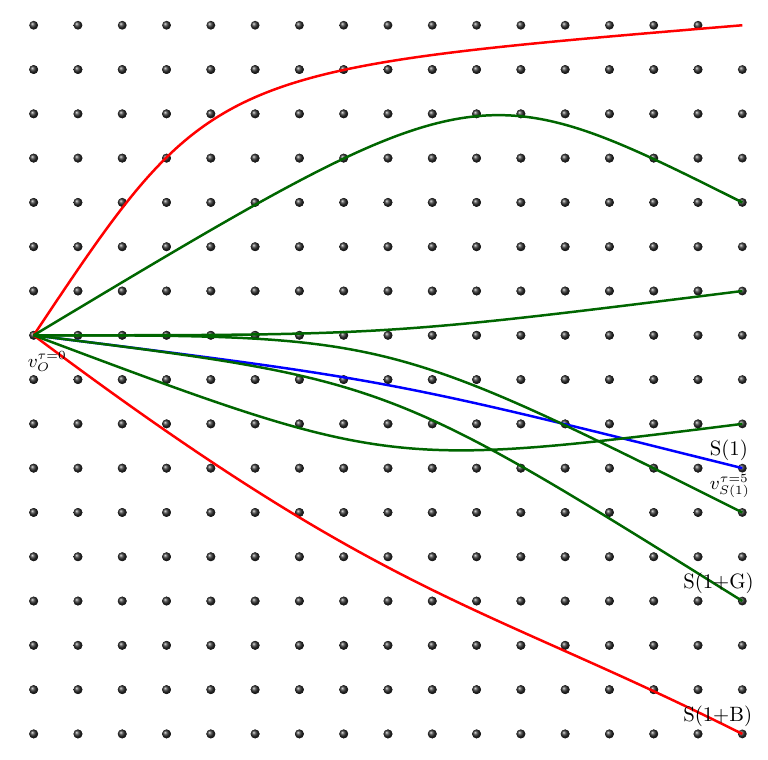}
    \caption{}
    \label{fig:lattice_vel}
  \end{subfigure}
  \caption{Representative trajectories based on the 3 sampling schemes over a $R^3$ lattice showing the spatial representation of the (a) path and (b) velocity profiles. Lattice points are connected with  cubic splines.}
  \label{fig:lattice}
\end{figure}

Each game is a $N$-player 2-level hierarchical game where level 1 actions for each agent are high level manoeuvres that are relevant to the task under execution, and level 2 actions are the corresponding trajectories. We setup the set of manoeuvres with the help of a first order logic rules (appendix \ref{appx:rules}) that takes into account the task of the vehicle and its situational state. The complete list of level 1 actions is documented in Table \ref{tab:l1_manvs}. Level 2 actions ($A_{i,2}$) are trajectories that are generated based on the actions in level 1. To generate the trajectories for each vehicle, we use a lattice sampling based trajectory generation similar to one presented in \citep{ziegler2009spatiotemporal}. First a set of lattice endpoints are sampled on $R^2$ cartesian co-ordinate centered on the vehicle's current position. Each lattice sample point on $R^2$ is then extended with a temporal lattice which is re-sampled to form the final lattice points in $R^3$ that contain the ($x,y$) positions and the target velocity at each lattice point after accounting for acceleration and jerk limits of passenger vehicles \citep{bae2019toward}. Finally, the sampled lattice points are connected with a smooth cubic spline that adhere to the velocity, acceleration, and jerk constraints, representing the vehicle trajectory (Fig. \ref{fig:lattice}). Appendix \ref{appx:traj_gen} explains this process in more detail.\par
Since the trajectory generation is in continuous space with infinite actions for the drivers to reason over, combined with the time constraints to make a decision (which is in the order of milliseconds), the situation is ripe for bounded rationality to be in play --- in a form where agents look at the game through samples\footnote{The usage of the word \emph{sample} here means an \emph{example}, and does not necessarily imply a presence of a specific statistical sampling procedure over a distribution.} of action rather than all possible trajectories over the continuous space. This form of bounded rationality is connected to Osborne and Rubinstein's model where agents' employ a mental process to sample other agents' actions and respond based on the imagined outcome of those samples \citep{osborne1998games}. In Osborne and Rubinstein's model, the model further develops this view of bounded rationality into an equilibrium (sampling equilibrium), whereas in our case we use this only as a method of action construction. We also use a common sampling procedure for all agents, and this enables the agents to have the same view of the action space and therefore play the same game. In the context of driving, this process is the same as when vehicles sample a set of trajectories of other agents and respond in accordance to those sampled trajectories with the additional assumption of having a common sampling scheme. Naturally, one may imagine that some sampling procedures make more sense than others, and in some procedures the assumption of a common sampling scheme is more reasonable than others. We now briefly mention the sampling procedures used in our experiments, and the intuitive reasoning behind each.\par
At each time step when the game tree is instantiated, agents observe the current attributes (such as position, velocity and acceleration) of other relevant agents in the game tree and use one of the following sampling methods to construct the game tree.
\begin{itemize}
    \item S(1): In the most basic case, an agent $i$ may sample a single trajectory (level 2 action) of every agent $-i$ that that they, i.e., $i$, think is most representative of the level 1 action of the agent they are currently reasoning over. To construct the trajectory sample, we select lattice endpoints along the lane centerline and use a piecewise constant acceleration model to generate the final trajectory.
    \item S(1+B): With a little more cognitive bandwidth, along with the $S(1)$ trajectory sample, they can also sample trajectories that form the extreme ends of the bounded level-2 action space of other agents. These trajectories are bounded spatially by the lane boundaries and temporally by the upper and lower bounds on the velocity limits of the level 1 action they correspond to. There are a total of 9 trajectory samples that are generated from this scheme; 3 velocity profiles generated over each of 3 path samples. This set of trajectories indicate what other agents might do in normative (i.e. following the rules as captured by the piecewise constant acceleration model) as well as in the extreme case but still within the physical limitations of the vehicle.
    \item S(1+G): The final sampling scheme lies in between the two schemes. Similar to $S(1+B)$, this scheme includes the $S(1)$ trajectory; however, the rest of the trajectories are sampled from a multivariate Gaussian distribution with $\mathbf{\mu} = [x_{S(1)},y_{S(1)},v_{S(1)}]^\top$ and an unit diagonal covariance matrix, where ($x_{S(1)},y_{S(1)},v_{S(1)}$) is the lattice endpoint corresponding to the $S(1)$ trajectory. We refer to this scheme as $S(1+G)$ and the samples include the normative behaviour that comes from $S(1)$ along with variations in the path and velocity of the vehicle but not to the extremes that were captured in the $S(1+B)$ scheme.
\end{itemize}
One can see that S(1) and S(1+B) are methods of action construction that do not \emph{sample} from a distribution in a statistical sense, and therefore the only assumption is that the agents share the general method of the action construction and the limits of the vehicle movement. In S(1+G) however, the specific samples of the distribution also need to be a part of the common knowledge for the agents to play the same game. In practice, this is a restrictive assumption; however, we include this method solely for comparison with other approaches.

\subsection{Utilities} 
\label{sec:one_shot:utils}
\begin{figure}[h]
\centering
\includegraphics[width=\columnwidth]{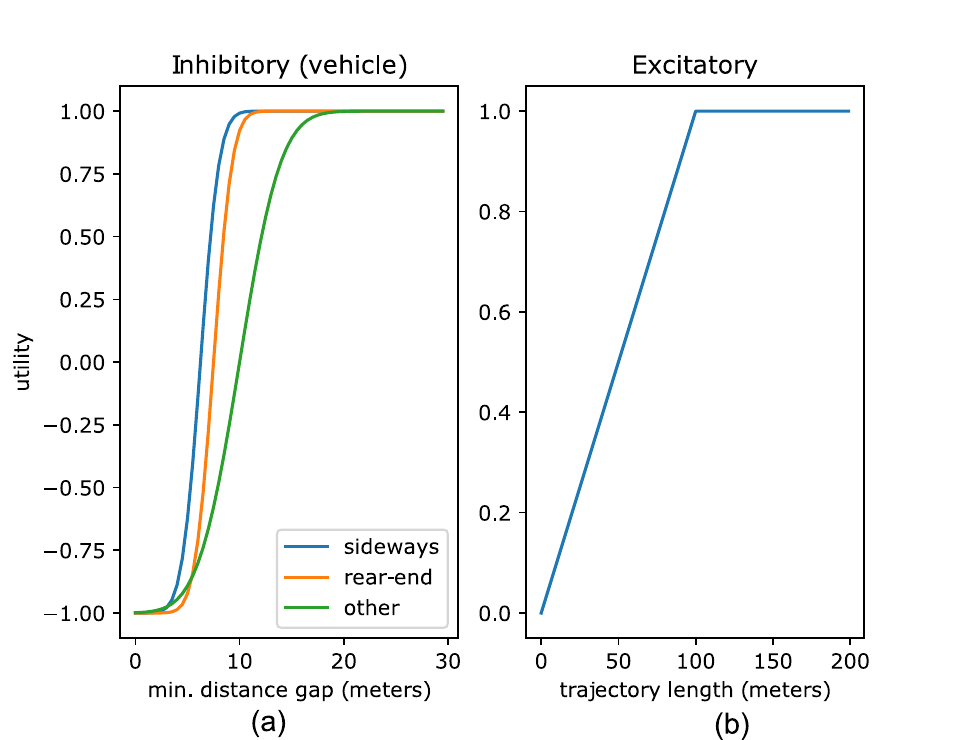}
  \caption{Utility function that maps a) minimum distance gap between trajectories to an utility interval [-1,1] (inhibitory utility for vehicle-vehicle interactions), and b) trajectory length to the utility interval [0,1] (excitatory utility).}
 \label{fig:one_shot_utils}
\end{figure}

To determine the utility structure, we draw from motivational aspects of driver behaviour modelling in traffic psychology literature \citep{summala1988risk}. In general, driving motivations are multiobjective and the broad dimensions can be classified into \emph{inhibitory} and \emph{excitatory} motives. Whereas excitatory motivations drive a driver to make progress towards reaching the destination, inhibitory motivations are the balancing factors that account for mitigating crashes and mental stress. The three different utlities used in this work are as follows
\begin{itemize}
    \item \textit{Excitatory utility}. the degree of progress a driver can make based on a selected trajectory $a_{i,2}$ is the excitatory utility $u_{\text{v\_exc}}(a_{i,2})$ as determined by the trajectory length $\left\lVert a_{i,2} \right\rVert$, $u_{\text{v\_exc}}(a_{i,2}) =  \min (\frac{\left\lVert a_{i,2} \right\rVert}{d_{g}},1)$, where $d_{g}$ is a constant and can be interpreted as the distance to goal or crossing the intersection. We set $d_{g} = 100$ (Fig. \ref{fig:one_shot_utils}b), which is approximately the distance to cross the intersection we study in our experiment. The minimum excitatory (i.e., progress) utility is set to 0 instead of -1 since even transiently waiting for another vehicle (trajectory length 0) may invoke a feeling of making some progress towards their destination.
    \item \textit{Vehicle inhibitory utility}. This utility, $u_{\text{v\_inh}}$, is based on the minimum distance gap between pairs of trajectories of vehicles. The form of the utility function is a sigmoidal function (Fig. \ref{fig:one_shot_utils}a), which are a popular family of functions that can map preferences over surrogate metrics, e.g., minimum distance gap $d(a_{i,2},a_{-i,2})$, into an utility interval \citep{fishburn1970utility}. For $u_{\text{v\_inh}}$, we first fix a minimum safe distance gap $d^{*}_{a_{i,2},a_{-i,2}}$ based on collision type that may occur as a result of the distance gap reaching zero between the trajectories of the agents in the game. Collision types include side collisions, rear-end collisions, and other types that include angle collisions. The value of the safe distance gap determines the location $\theta$ of the sigmoidal function ($\erf$). However, since the conception of what is considered safe may vary in a population of drivers, we let $\theta$ to be a random variable that is normally distributed with $\mu=d^{*}_{a_{i,2},a_{-i,2}}$ and constant variance $\sigma$ determining the scale of the sigmoidal function. The choice of $\erf$ as the sigmoidal function is a mathematical convenience since the Gaussian integral of the $\erf$ in $u_{\text{v\_inh}}(a_{i,2},a_{-i,2})$ evaluates to another sigmoidal $\erf(\frac{d(a_{i,2},a_{-i,2})-d^{*}_{a_{i,2},a_{-i,2}}}{2\sigma})$. 
    \begin{align}
        u_{\text{v\_inh}}(a_{i,2},a_{-i,2}) =& \nonumber \\ \int \erf\left[\frac{d(a_{i,2},a_{-i,2})-\theta}{\sigma\sqrt{2}}\right]\mathcal{N}&(\theta;d^{*}_{a_{i,2},a_{-i,2}},\sigma) d\theta
    \end{align}
    \item \textit{Pedestrian inhibitory utility}. This utility, $u_{\text{p\_inh}}$, is a step function over [-1,1] such that $u_{\text{p\_inh}}(a_{i,2}) = -1$ if $a_{i,2}$ is a trajectory that does not wait for a pedestrian when the pedestrian is in the vicinity having a right of way, or is on the crosswalk to be traversed; and 1 otherwise.
\end{itemize}
The above multi objective utilities are aggregated using a weighted aggregation method with a weight vector $\mathbf{W}$ that combines inhibitory and excitatory utilities to produce a single real value.
\begin{align}
\scriptsize
u_{i}(a_{i,2},a_{-i,2}) =  \mathbf{W} \cdot \begin{bmatrix}
u_{\text{v\_inh}}(a_{i,2},a_{-i,2}) & u_{\text{p\_inh}}(a_{i,2}) & u_{\text{v\_exc}}(a_{i,2})
\end{bmatrix}^\top
\end{align}
\normalsize
The utilities for the actions in $\mathcal{G}_1$ depends on the solution concept (discussed in the next section), and can be calculated as follows. $u_{i}(a_{i,1},a_{-i,1}) = V_{2,\eta}^{*}(i)$, where $\eta$ is the leaf node of the branch $a_{i,1},a_{-i,1}$ and $V_{2,\eta}^{*}(i)$ is the utility of agent $i$ following the pure strategy response $a_{i,2}^{*}$, where $a^{*}_{i,2}$ is the solution to the underlying $\mathcal{G}_2$ game. The utilities presented above are the utilities at the leaf nodes of the game (as represented by $U$ based on the formalization in Sec. \ref{ch:one_shot:formalization}).
\section{Solution concepts in hierarchical games}
\label{sec:soln_concept}
A key element that influences solution concepts in games is the manner in which each agent reasons over the strategies of other agents. In non-strategic behaviour models, agents do not explicitly model other agents in the game and respond solely on the basis of their own utility structure. Wright and Leyton-Brown \citep{wright2020formal} refer to this property of behaviour being dependent only on the agent's own payoff as \emph{dominance responsive}. Along with following the property of dominance responsiveness, strategic agents, on the other hand, also reason over the strategies of other agents. This means that strategic agents are not only responsive to their own utilities, but also demonstrate a behaviour that is dependent on others' utilities too; a property that is referred to as \emph{other responsiveness} \citep{wright2020formal}. \par
The first category of behaviour models we consider is the Quantal level-k (Qlk) model \citep{wright2012behavioural}. Qlk models the population of agents as a mix of strategic and non-strategic agents, with agents having a bounded iterated cognitive hierarchy of reasoning. Strategic agents in Qlk use Quantal Best Response (QBR) function, often expressed as a logit response $\pi_{i}^{\text{QBR}}({a_{i},s_{-i},\lambda)} = \frac{\exp{[\lambda \cdot u_{i}(a_{i},s_{-i})]}}{\Sigma_{a_{i}^{'}}\exp{[\lambda \cdot u_{i}(a_{i}^{'},s_{-i})]}}$, where $s_{-i}$ represents the pure or mixed strategies of other agents and $\lambda$ is the \emph{precision} parameter that can account for errors in agent response with respect to utility differences\footnote{In this formulation, the symbols $s_{i}$ and $a_{i}$ are strategies and actions of a game in a general sense, respectively, and not related to the symbols used specifically in the formulation of hierarchical games earlier.}. When $\lambda \rightarrow 0$, the mixed response is a uniform random distribution, whereas $\lambda \rightarrow \infty$ makes the response equivalent to best response. Level-0 agents are non-strategic (NS) agents who choose their actions uniformly at random, whereas Level-1 agents are strategic (S) agents who believe that the population consists solely of Level-0 agents, and their response is a QBR response to Level-0 agents' actions. 
In the original Qlk model, level-0 agents follow a mixed uniform distribution strategy; however, in \citep{wright2014level}, the behaviour of level-0 agents is formalised by noting that any behaviour that is \emph{dominance responsive only} can be considered a model of level-0 behaviour. Additionally, the behaviour that is \emph{other responsive}, are level-$k \geqslant 1$ behaviour. In this  paper, we use this expanded definition, and similar to \citep{wright2014level}, the level-0 agents' strategies follow more intuitive, yet nonstrategic response, such as the maxmax response (MX) or the maxmin (MM) response. Note that for both models, an agent only needs to perform operations on their own utilities, thus adhering to the level-0 constraint of not being \emph{other responsive}.
We believe that the expanded definition of the level-0 agents suits our situation much better, since it is unrealistic to expect a driver to choose actions purely at random from their available actions. Even with this expanded definition, these are still \emph{non-strategic} since level-0 agent responses depend purely on their own utilities and do not rely on strategic reasoning over other agents' utilities \citep{wright2020formal}.\par
Another category of non-strategic behaviour that we consider is rule following behaviour. Under a rule following behaviour level-0 agents strictly adhere to the traffic rules regardless of what the utilities may suggest. Such a strict rule following behaviour is not even \emph{dominance responsive} because based on the utilities constructed in Section \ref{sec:one_shot:utils}, the action that the rule suggests can be strictly dominated by all other actions, and the agent would still follow the rule. On the other hand, if the utilities are constructed in an alternate way that captures the preference of the rule following, then our level-0 agent can be deemed to be dominance responsive. Based on these characteristics, the rule following can be said to be a nonstrategic behaviour, therefore in the category of level-0.\par
In a hierarchical game, since the agent strategies are factored into levels $s_{i} = (a_{i,1},a_{i,2})$, the manner in which an agent reasons over strategies in one level might not be the same as the reasoning process in another level. Therefore, instead of a single solution concept in the game of Fig. \ref{fig:static_hier}, action level games $\mathcal{G}_{2}$ can have a different solution concept than the one in the game $\mathcal{G}_{1}$. In our models, we let agents have a cognitively less demanding non-strategic response in $\mathcal{G}_{2}$, and a more deliberative strategic response in $\mathcal{G}_{1}$. This choice is similar to one taken in \citep{fisac2019hierarchical}, and reflects the natural process where it is easier for drivers to reason strategically over the strategy space of discrete manoeuvres than over the space of infinitely many trajectories.\par
We consider three metamodels of behaviour under Qlk: Ql0, Ql1, and QlkR. We refer to them as metamodels, since they can be further refined based on the choice of response function and sampling schemes to create concrete models (see Table \ref{ch:oneshot:all_behaviourmodel_table}).\par

\noindent \textit{Ql0 metamodel.} In QL0 metamodel, we restrict the population to be solely level-0 responders in both $\mathcal{G}_1$ and $\mathcal{G}_2$.
\begin{table}[t]
\centering
\resizebox{0.8\columnwidth}{!}{%
\begin{tabular}{cccccccc}
\toprule
\multicolumn{2}{c}{QL0} & \multicolumn{2}{c}{QL1} & \multicolumn{2}{c}{PNE-QE}& \multicolumn{2}{c}{QlkR}\\
\cmidrule(lr{1em}){1-2}\cmidrule(lr{1em}){3-4}\cmidrule(lr{1em}){5-6}\cmidrule(lr{1em}){7-8}
$\mathcal{G}_{1}$&$\mathcal{G}_{2}$&$\mathcal{G}_{1}$&$\mathcal{G}_{2}$&$\mathcal{G}_{1}$&$\mathcal{G}_{2}$&$\mathcal{G}_{1}$&$\mathcal{G}_{2}$\\
\midrule
NS&NS&S+NS&NS&S&NS&S&NS\\
\bottomrule
\end{tabular}}
\caption{Distribution of strategic (S) and non-strategic (NS) behaviour in action level games $\mathcal{G}_1$ and $\mathcal{G}_2$ in four metamodels QL0, QL1, PNE-QE, and QlkR.}
\label{tab:s_ns}
\end{table}
Level-0 agents follow non-strategic behaviour with one of two solution concepts, maxmax response (MX), and best worst-case or maxmin response (MM). The model of MX is:
{\small
\begin{align}
    a_{i,\kappa}^{*} &= \argmaxB_{\forall a_{i,\kappa},a_{-i,\kappa}} u_i(a_{i,\kappa},a_{-i,\kappa}) \label{ns_br} \\
    \pi_{i}(a_{i,\kappa})  &= \frac{\exp[\lambda_{i} \cdot u_{i}(a_{i,\kappa},\argmaxB_{\forall a_{-i,\kappa}} u_i(a_{i,\kappa},a_{-i,\kappa}))]}{\Sigma_{\forall a_{i,\kappa}} \exp[\lambda_{i} \cdot u_{i}(a_{i,\kappa},\argmaxB_{\forall a_{-i,\kappa}} u_i(a_{i,\kappa},a_{-i,\kappa}))]}  \label{ns_mbr}
\end{align}
}
where $a_{i}^{*}$ is the pure-strategy utility-maximizing action for $i$. The model for nonstrategic MM response is:
{\small
\begin{align}
    a_{i,\kappa}^{*} &= \argmaxB_{\forall a_{i,\kappa}} \argminB_{\forall a_{-i,\kappa}} u_i(a_{i,\kappa},a_{-i,\kappa}) \label{ns_mm} \\
    \pi_{i}(a_{i,\kappa}) &= \frac{\exp[\lambda_{i} \cdot u_{i}(a_{i,\kappa},\argminB_{\forall a_{-i,\kappa}} u_i(a_{i,\kappa},a_{-i,\kappa}))]}{\Sigma_{\forall a_{i,\kappa}} \exp[\lambda_{i} \cdot u_{i}(a_{i,\kappa},\argminB_{\forall a_{-i,\kappa}} u_i(a_{i,\kappa},a_{-i,\kappa}))]}  \label{ns_mmm}
\end{align}
}
Equations \ref{ns_mbr} and \ref{ns_mmm} are relaxations that translate the pure strategy action to a noisy response $\pi_i(a_{i,\kappa})$ based on the precision parameter $\lambda_{i}$ and sensitivity to $i$'s utility difference with respect to opponent actions that maximizes $i$'s utility for MX and minimizes for MM.\par
\noindent \textit{Ql1 metamodel.} In QL1, the population consists of a mix of level-0 and level-1 responders in $\mathcal{G}_1$ and level-0 responders in $\mathcal{G}_2$ (Table \ref{tab:s_ns}). Level-0 agents in this population follow MX and MM nonstrategic behaviour as formulated earlier, and level-1 agents best respond quantaly to level-0 agents' behaviour. With the expanded definition of level-0 agents as nonstrategic bounded rational agents, there is a design choice to be made on what level-1 agents believe about level-0 agents. They can consider level-0 agents to be bounded rational responders having mixed response of Equations \ref{ns_mbr} and \ref{ns_mmm}, or level-1 agents can consider level-0 agents to be pure strategy rational responders based on Equations \ref{ns_br} and \ref{ns_mm}. We choose the latter to align with the original Qlk model, where agents modelling other agents as bounded rational agents are observed only at a higher cognitive level (level-2 and above). In Qlk models, the mixed population is modelled as a bimodal mixture distribution. Therefore, if the proportion of level-0 and level-1 agents is $\alpha$ and $1-\alpha$, respectively, then the QL1 model response in $\mathcal{G}_1$ is the mixed strategy response.
\begin{equation}
\pi_{i}^{\text{QL1}}(a_{i,1}) = \alpha \cdot \pi_{i}^{\text{QL0}}(a_{i,1}) + (1-\alpha) \cdot \pi_{i}^{\text{QBR}}({a_{i,1},a^{*}_{-i,1},\lambda_{i})}
\label{eqn:ql1_g1}
\end{equation}
where $\pi_{i}^{\text{QL0}}(a_{i,1})$ is the left hand side of the equation \ref{ns_mbr} or \ref{ns_mmm} and $a^{*}_{-i,1}$ is the solution set to equations \ref{ns_br} or \ref{ns_mm} for each of the other agents. \par

\noindent \textit{QlkR metamodel.} In the QlkR metamodel, the population consists of level-1 agents who believe that everyone else follows a rule following, and the agents in the QlkR model best respond quantaly with precision parameter $\lambda_{i}$. The table of rules that determine the behaviour of rule-following is included in Appendix \ref{appx:rules}. The rule following behaviour in this case can be considered as an alternate model of level-0 behaviour. However, the only property that a level-0 model adheres to is dominance responsive. One can argue that such a strict rule following behaviour is not even \emph{dominance responsive} because based on the utilities constructed in Section \ref{sec:one_shot:utils}, the action that the rule suggests can be strictly dominated by all other actions, and the agent would still follow the rule. On the other hand, if the utilities are constructed in an alternate way that captures the preference of the rule following, then our level-0 agent can be deemed to be dominance responsive. Let $\mathcal{R}_{-i}(X_{-i})$ be the action corresponding to the traffic rule that agent $-i$ should follow in state $X_{-i}$, then the model of QlkR behaviour is as follows.\par
\begin{align}
 a_{i,\kappa}^{*} &= \argmaxB_{\forall a_{i,\kappa},\mathcal{R}_{-i}(X_{-i})} u_i(a_{i,\kappa},\mathcal{R}_{-i}(X_{-i})) \label{qlkr_model_eqn_br} \\
    \pi_{i}^{\text{QlkR}}(a_{i,\kappa})  &= \frac{\exp[\lambda_{i} \cdot u_{i}(a_{i,\kappa}, u_i(a_{i,\kappa},\mathcal{R}_{-i}(X_{-i}))]}{\Sigma_{\forall a_{i,\kappa}} \exp[\lambda_{i} \cdot u_{i}(a_{i,\kappa},\mathcal{R}_{-i}(X_{-i}))]}  
    \label{qlkr_model_eqn}
\end{align}
\noindent \textit{PNE metamodel.} The final metamodel we consider is a generalization of pure strategy Nash equilibrium with noisy response. In this metamodel, agents follow a non-strategic model in $\mathcal{G}_2$, and a strategic model in $\mathcal{G}_1$ as described below.
{\small
\begin{equation}
    a_{i,1}^{*} = \argmaxB_{\forall a_{i,1}} u_i(a_{i,1},a_{-i,1}^{*}) \label{ns_ne}
    \end{equation}
    \begin{align}
    &\pi_{i}^{\text{PNE-QE}}(a_{i,1})   = \nonumber \\ & \frac{\exp[-\lambda_{i} \cdot \min_{\forall (a_{i,1}^{*},a_{-i,1}^{*})}( u_{i}^{*} - u_i(a_{i,1},a_{-i,1}^*))]}{\Sigma_{\forall a_{i,1}} \exp[-\lambda_{i} \cdot \min_{\forall (a_{i,1}^{*},a_{-i,1}^{*})}( u_{i}^{*} - u_i(a_{i,1},a_{-i,1}^*))]}  \label{ns_mne}
\end{align}
}
where $u_{i}^{*} = u_{i}(a_{i,1}^{*},a_{-i,1}^{*})$. In the above model, agents respond according to pure strategy Nash equilibria $a_{i,1}^{*}$, but in error may choose actions $a_{i,1} \notin a_{i,1}^{*}$ based on the sensitivity to the difference in the utility of the action and an equilibrium action. We refer to this model as pure strategy Nash equilibria with quantal errors (PNE-QE). The formulation is similar to Quantal Response Equilibrium (QRE), yet with key differences. In QRE, strategic reasoning occurs in a space of mixed responses and the precision parameter is part of common knowledge in the game. In our model, reasoning over opponent strategies is in pure strategy action space and the precision parameter is endogenous to each agent; therefore, when an agent reasons about the strategies of other agents, their parameters do not play a role \citep{crawford2013structural}.
Based on the choice of the metamodel, the response function, and the sampling scheme, we get 30 different behaviour models ($\mathcal{B}$), cf. Table \ref{ch:oneshot:all_behaviourmodel_table}, which we evaluate in the next section. \par
\subsubsection{Estimation of game parameters} The dataset used in this  paper includes $\mathcal{D}$ (\textasciitilde 23k) hierarchical games, instantiated with planning frequency $\Delta t_{p}=1s$, and planning horizon $\Delta t_{h}=5s$ and with the state variables $X_{i}$ along with the observed strategy $s_{i}^{o}=(a_{i,1}^{o},a_{i,2}^{o})$ for every agent $i$ in the game. For each behaviour model $b \in \mathcal{B}$, we note the errors in actions with respect to the pure strategy responses in the games as $\Delta\mathcal{U}_{b} = \{\epsilon_{i,b}|\epsilon_{i,b} = \min_{\forall a_{i}^{*}}[u_{i}(a_{i}^{*},a_{-i}^{*}) - u_{i}(a_{i}^{o},a_{-i}^{*})]\}$, where $a_{i}^{*}$ are the solutions to Equations \ref{ns_br} or \ref{ns_mm} for non-strategic models, Equation \ref{qlkr_model_eqn_br} for QlkR model, and \ref{ns_ne} for PNE-QE model (we verified the existence of pure strategy NE for all $\mathcal{G}_1$ games in $\mathcal{D}$). Within the context of a game, we assume that all players follow a common behaviour model, and the precision parameters ($\lambda_{i,b}$) in an individual game is a function of the agent's state $X_{i}$ (see Tables \ref{tab:x_l_table1} and \ref{tab:x_l_table2} for the list of state factors) from whose perspective the game is initiated as well as the behaviour model $b$ of the game. Therefore, for a given state factor $X_{i}$, $\epsilon_{i,b}$, or the error value that captures the utility difference follows an exponential distribution based on the game's precision parameter for Ql0, QlkR and PNE-QE metamodels, and a mixed exponential distribution (\ref{eqn:ql1_g1}) for QL1 in $\mathcal{G}_{1}$. The exponential distribution of the errors and the (assumed) dependency based on state factors lend well for the model to be fit based on a \emph{generalized linear model} \citep{gelman2006data}. Additionally, since the mean of an exponential distribution is just the inverse of the distribution parameter, the estimate that the \emph{glm} model gives is the inverse of the precision parameter estimate that we wish to infer. Therefore, to estimate the value of  $\lambda_{i,b}$ we fit a generalized linear model $glm(\epsilon_{i,b} \sim \bm{\beta}X_{i})|_{\Delta \mathcal{U}_{b}}$ with Gamma($k=1$) family and inverse link, which models $\epsilon_{i,b}$ as an exponentially distributed random variable with $\text{E}[\epsilon_{i,b}] = \frac{1}{\lambda_{i,b}}$ and $\text{Var}[\epsilon_{i,b}] = \frac{1}{\lambda_{i,b}^2}$. $\bm{\beta}$ is the model co-efficient, solved through maximum likelihood estimate based on the data in $\Delta \mathcal{U}_{b}$. The prediction of the $glm$ model gives the mean and standard error of $\lambda_{i,b}^{-1}$ based on the state observation $X_{i}$. For the mixed exponential distribution in QL1 model, once we estimate the individual precision parameters of \ref{eqn:ql1_g1}, we estimate $\alpha$ by maximizing the likelihood function $\Sigma_{\forall a_{i,1}^{o}}\ln(\pi^{\text{QL1}}_{i}(a_{i,1}^{o}))$.
\begin{table*}[ht]
\centering
\small
\begin{tabular}{@{}lp{2cm}p{3cm}p{2cm}l@{}}
\toprule
\textbf{Model name}  & \textbf{Metamodel} & \textbf{Action level game $\mathcal{G}_{1}$} & \textbf{Action level game $\mathcal{G}_{2}$} & \textbf{Trajectory sampling} \\ \midrule
PNE-QE:MM\_S(1+B)    & PNE   & Pure strategy NE                   & Maxmin                                              & Bounds, S(1+B)                         \\
PNE-QE:MM\_S(1+G)    & PNE   & Pure strategy NE                   & Maxmin                                              & Truncated Gaussian, S(1+G)             \\
PNE-QE:MX\_S(1+B)    & PNE   & Pure strategy NE                   & Maxmax                                              & Bounds, S(1+B)                         \\
PNE-QE:MX\_S(1+G)    & PNE   & Pure strategy NE                   & Maxmax                                              & Truncated Gaussian, S(1+G)             \\
PNE-QE\_S(1)         & PNE   & Pure strategy NE                   & NA                                                  & Prototype trajectory, S(1)             \\
Ql0:MM:MM\_S(1+B)    & Ql0                & Maxmin                                           & Maxmin                                              & Bounds, S(1+B)                         \\
Ql0:MM:MM\_S(1+G)    & Ql0                & Maxmin                                           & Maxmin                                              & Truncated Gaussian, S(1+G)             \\
Ql0:MM:MX\_S(1+B)    & Ql0                & Maxmin                                           & Maxmax                                              & Bounds, S(1+B)                         \\
Ql0:MM:MX\_S(1+G)    & Ql0                & Maxmin                                           & Maxmax                                              & Truncated Gaussian, S(1+G)             \\
Ql0:MM\_S(1)         & Ql0                & Maxmin                                           & NA                                                  & Prototype trajectory, S(1)             \\
Ql0:MX:MM\_S(1+B)    & Ql0                & Maxmax                                           & Maxmin                                              & Bounds, S(1+B)                         \\
Ql0:MX:MM\_S(1+G)    & Ql0                & Maxmax                                           & Maxmin                                              & Truncated Gaussian, S(1+G)             \\
Ql0:MX:MX\_S(1+B)    & Ql0                & Maxmax                                           & Maxmax                                              & Bounds, S(1+B)                         \\
Ql0:MX:MX\_S(1+G)    & Ql0                & Maxmax                                           & Maxmax                                              & Truncated Gaussian, S(1+G)             \\
Ql0:MX\_S(1)         & Ql0                & Maxmax                                           & NA                                                  & Prototype trajectory, S(1)             \\
Ql1:MM:MM\_S(1+B)    & Ql1                & BR to Maxmin                 & Maxmin                                              & Bounds, S(1+B)                         \\
Ql1:MM:MM\_S(1+G)    & Ql1                & BR to Maxmin                 & Maxmin                                              & Truncated Gaussian, S(1+G)             \\
Ql1:MM:MX\_S(1+B)    & Ql1                & BR to Maxmin                 & Maxmax                                              & Bounds, S(1+B)                         \\
Ql1:MM:MX\_S(1+G)    & Ql1                & BR to Maxmin                 & Maxmax                                              & Truncated Gaussian, S(1+G)             \\
Ql1:MM\_S(1)         & Ql1                & BR to Maxmin                  & NA                                                  & Prototype trajectory, S(1)             \\
Ql1:MX:MM\_S(1+B)    & Ql1                & BR to Maxmax                  & Maxmin                                              & Bounds, S(1+B)                         \\
Ql1:MX:MM\_S(1+G)    & Ql1                & BR to Maxmax                  & Maxmin                                              & Truncated Gaussian, S(1+G)             \\
Ql1:MX:MX\_S(1+B)    & Ql1                & BR to Maxmax                  & Maxmax                                              & Bounds, S(1+B)                         \\
Ql1:MX:MX\_S(1+G)    & Ql1                & BR to Maxmax                  & Maxmax                                              & Truncated Gaussian, S(1+G)             \\
Ql1:MX\_S(1)         & Ql1                & BR to Maxmax                  & NA                                                  & Prototype trajectory, S(1)             \\
QlkR:BR-R:MM\_S(1+B) & QlkR                & BR to traffic rule                    & Maxmin                                              & Bounds, S(1+B)                         \\
QlkR:BR-R:MM\_S(1+G) & QlkR                & BR to traffic rule                    & Maxmin                                              & Truncated Gaussian, S(1+G)             \\
QlkR:BR-R:MX\_S(1+B) & QlkR                & BR to traffic rule                    & Maxmax                                              & Bounds, S(1+B)                         \\
QlkR:BR-R:MX\_S(1+G) & QlkR                & BR to traffic rule                    & Maxmax                                              & Prototype trajectory, S(1)             \\
QlkR:MX\_S(1)        & QlkR                & BR to traffic rule                    & NA                                                  & Prototype trajectory, S(1)             \\ \bottomrule
\end{tabular}
\caption{Synopsis of the thirty behaviour models included in the evaluation. 'BR' stands for Best response.}
\label{ch:oneshot:all_behaviourmodel_table}
\end{table*}
\normalsize
\section{Experiment and evaluation}
\label{sec:experiments}
\noindent \textbf{Dataset.} We used the intersection dataset of Waterloo Multi-Agent Traffic Dataset, which contains a total of 3649 vehicles and 264 pedestrians, including their centimetre-accurate trajectory estimates. We analyse the decision making in right turning and left turning vehicles, which results in a total of 12526 hierarchical games. Table \ref{tab:l1_manvs} shows the manoeuvres that are used in the construction of the level-1 games. The manoeuvres are context specific, and we use a rule based method (appendix \ref{appx:rules}) to generate the set of available manoeuvres to each agent in the game. The situational context in which each maneuver is available to an agent is shown in the description column of Table \ref{tab:l1_manvs}. Relevant code for the experiments is available at \url{https://git.uwaterloo.ca/a9sarkar/traffic_behaviour_modeling} and dataset is available at \url{https://uwaterloo.ca/waterloo-intelligent-systems-engineering-lab/datasets/waterloo-multi-agent-traffic-dataset-intersection}. \par
\begin{table*}[!h]
\centering
\resizebox{0.8\textwidth}{!}{%
\begin{tabular}{lp{8cm}}
\toprule
Level-1 action (maneuver) & Description\\
\midrule
\texttt{wait-for-oncoming (aggressive)}&\multirow{4}{8cm}{Applies to left and right turning vehicles. The action of waiting for a vehicle that has the right of way. Generates a trajectory with terminal velocity of zero.}\\
\texttt{wait-for-oncoming (normal)}\\
&\\
&\\
\midrule
\texttt{proceed-turn (aggressive)}&\multirow{2}{8cm}{Applies to left and right turning vehicles. Action of executing the turn.}\\
\texttt{proceed-turn (normal)}\\
\midrule
\texttt{track-speed (aggressive)}&\multirow{2}{8cm}{Applies to straight through vehicles. Trajectory accelerates or decelerates to road speed limit.}\\
\texttt{track-speed (normal)}\\
\midrule
\texttt{follow-lead (aggressive)}&\multirow{3}{8cm}{Applies to straight through vehicles vehicles with a lead vehicle. Generates a trajectory with same target velocity as leading vehicle.}\\
&\\
\texttt{follow-lead (normal)}\\
\midrule
\texttt{decelerate-to-stop (aggressive)}&\multirow{4}{8cm}{Applies to all vehicles. Indicates vehicles coming to a stop on change of traffic light from green to amber/red. Generates a trajectory with terminal velocity of zero.}\\
&\\
&\\
\texttt{decelerate-to-stop (normal)}\\
\midrule
\texttt{wait-for-lead-to-cross (aggressive)}&\multirow{4}{9cm}{Applies to left and right turning vehicles with a lead vehicle. Indicates vehicle waiting for a lead vehicle to finish executing its turn. Generates a trajectory with terminal velocity of zero.}\\
&\\
&\\
\texttt{wait-for-lead-to-cross (normal)}&\\
\midrule
\texttt{follow-lead-into-intersection (aggressive)}&\multirow{5}{8cm}{Applies to left and right turning vehicles with a lead vehicle which is yet to execute the turn. Indicates a vehicle following the its vehicle into the intersection while the lead vehicle executes a turn. Generates a trajectory with same target velocity as leading vehicle.}\\
&\\
&\\
&\\
\texttt{follow-lead-into-intersection (normal)}\\
\midrule
\texttt{wait-on-red (aggressive)}&\multirow{2}{8cm}{Applies to all vehicles. Indicates vehicles waiting on red light.}\\
\texttt{wait-on-red (normal)}\\
\midrule
\texttt{wait-for-pedestrian (aggressive)}&\multirow{2}{9cm}{Applies to left and right turning vehicles. Indicates waiting for a pedestrian to cross a crosswalk.}\\
\texttt{wait-for-pedestrian (normal)}&\\
\bottomrule
\end{tabular}}
\caption{Description of actions used in $\mathcal{G}_1$ of the hierarchical game. \texttt{aggressive} actions generate trajectories with maximum absolute acceleration/deceleration $\ge 2\,\text{ms}^{-2}$.}
\label{tab:l1_manvs}
\end{table*}
In this experiment we study naturalistic driving behaviour and evaluate which behaviour model captures human driving better, both in terms of model fit and predictive accuracy. We set $\mathbf{W} = \begin{bmatrix}
0.25 & 0.5 & 0.25\end{bmatrix}$, thereby giving more importance to pedestrian inhibitory actions and set the value of $d_{g}=100\,$m. In particular we answer the following research questions:
\begin{itemize}
    \item \emph{RQ1}. Which solution concept provides the best explanation for the observed naturalistic data?
    \item \emph{RQ2}. How do state factors influence the precision parameters in the games?
    \item \emph{RQ3}. How does the choice of the response function in the lower action level game $\mathcal{G}_2$ affect the higher level solutions in $\mathcal{G}_1$?
\end{itemize}
%


\begin{figure*}[ht]%
\centering
\includegraphics[width=\textwidth]{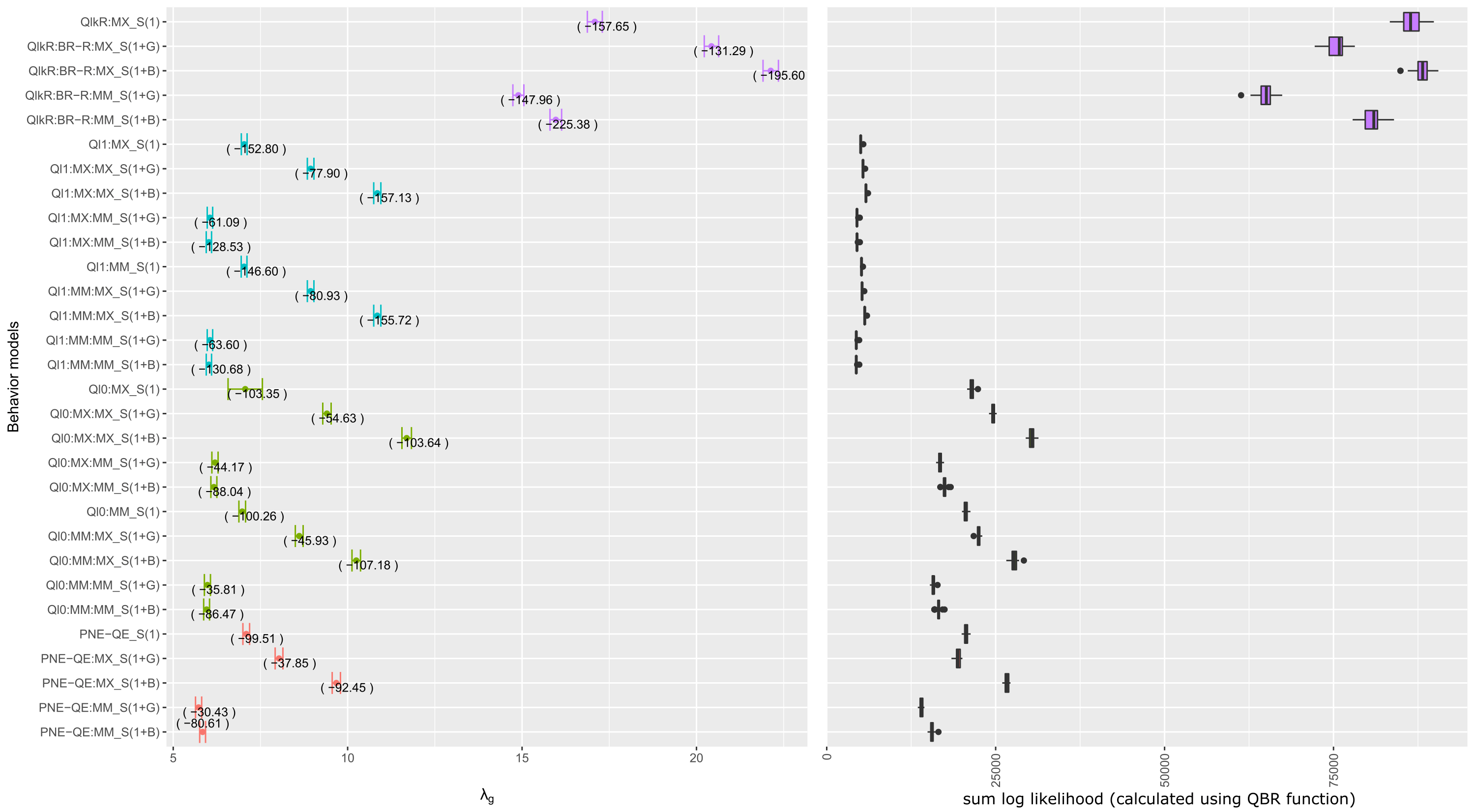}%
\caption{Comparison of models based on (a) precision parameters ($\lambda_{i,b}$), fit (AIC values marked in brackets), and (b) predictive accuracy (sum log likelihood calculated from QBR function of observations in test data after 30 runs). The first plot shows the mean estimate along with the standard error of the precision parameter across every state, and the second plot shows the boxplot of the likelihood estimates across 30 runs.}
\label{fig:lambda_chart}
\end{figure*}

\begin{figure*}[ht]%
\centering
\includegraphics[width=0.8\textwidth]{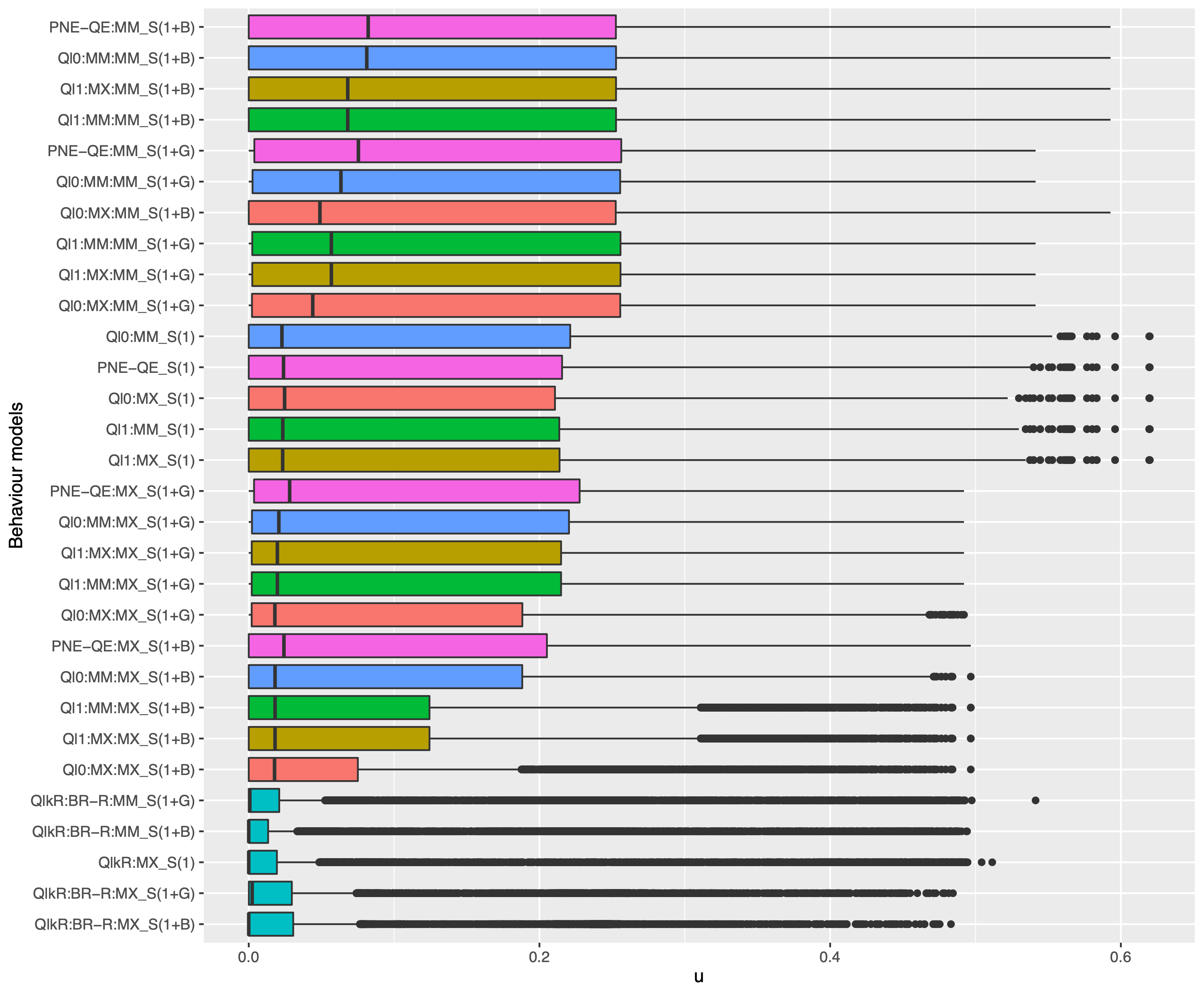}%
\caption{Comparison of the models based on spread of utility differences ($\Delta u$) between selected action ($a^{o}_{i,1}$) and the solution ($a^{*}_{i,1}$) in $\mathcal{G}_{1}$ games for each behaviour model, sorted by mean $\Delta u$. }
\label{fig:delta_u_distr}
\end{figure*}


Table \ref{ch:oneshot:all_behaviourmodel_table} shows a synopsis of all the behaviour models included in the evaluation. The model names are indexed by their metamodel followed by the choices of the response functions in $\mathcal{G}_1$:$\mathcal{G}_2$ followed by the sampling scheme used in $\mathcal{G}_2$. For QlkR metamodels, recall that level-1 agents believe that other agents will follow the traffic rules. Therefore, level-1 agents when solving their own action level game $\mathcal{G}_2$, would only solve the game under the manoeuvre that corresponds to the rule following behaviour on the part of level-0 agents. The MX or MM solution concept noted for QlkR metamodels is the one used to solve the $\mathcal{G}_2$ games.  For models using S(1) sampling of trajectories, the response function in $\mathcal{G}_2$ is omitted, since the hierarchical game consists only of $\mathcal{G}_1$ games; and in those cases, each agent has a single choice under each $\mathcal{G}_2$ roots. We perform our analysis of RQs 1 and 2 based on $\mathcal{G}_1$, and discuss the impact of the choice of $\mathcal{G}_2$ solution concepts as part of RQ3. The list of manoeuvres for $\mathcal{G}_1$ is shown in Table \ref{tab:l1_manvs} along with their descriptions. Each manoeuvre is further divided into aggressive and normal modes, thus giving a total of 18 $\mathcal{G}_1$ actions.\par
\emph{RQ1. Which solution concept provides the best explanation for the observed naturalistic data? } We address this question in three ways; with respect to (i) parameter values in the model, (ii) predictive accuracy in unseen data, and (iii) model fit.
Fig. \ref{fig:delta_u_distr} shows the box plot of $\Delta \mathcal{U}_{b}$ or the utility difference between the true utility (i.e. utility of observed manoeuvre) and the utility of the manoeuvre predicted by the game solutions for each behaviour model. Therefore, a lower value indicates that the solutions are closer to the true manoeuvre executed by the vehicle. For cases where there are multiple solutions, the one with the minimum $\Delta \mathcal{U}_{b}$ is chosen. The models in the figure are sorted based on the mean $\Delta \mathcal{U}_{b}$. We see that the QlkR metamodel consistently shows lower values in the utility difference compared to the Ql0, Ql1, and PNE-QE metamodels. Within the QlkR metamodels, utilities of the actions selected by the QlkR:MM\textsubscript{S(1+B)} model are closest to the utilities of real action selected by the vehicles, however, the difference among the QlkR models is not as distinct.\par
\begin{table}[t]
\small
\centering
\begin{tabular}{cP{.6cm}P{.6cm}P{.6cm}P{.6cm}}
    \toprule
    $\mathcal{R}(\mathcal{G}_1)$:&\multicolumn{2}{c}{MX}&\multicolumn{2}{c}{MM}\\
    \cmidrule(lr{1em}){2-3}\cmidrule(lr{1em}){4-5}
    $\mathcal{R}(\mathcal{G}_2)$:&MX&MM&MX&MM\\
    \midrule
    QL0&+5.5&--&+4.5&--\\
    QL1&+4.9&--&+4.9&--\\
    \bottomrule
\end{tabular}
\hspace{1cm}
\begin{tabular}{cP{.6cm}P{.6cm}P{.6cm}P{.6cm}}
    \toprule
    $\mathcal{R}(\mathcal{G}_1)$:&\multicolumn{2}{c}{PNE-QE}&\multicolumn{2}{c}{QlkR}\\
    \cmidrule(lr{1em}){2-3}\cmidrule(lr{1em}){4-5}
    $\mathcal{R}(\mathcal{G}_2)$:&MX&MM&MX&MM\\
    \midrule
    &+3.9&--&+6.3&--\\
    \bottomrule
\end{tabular}
\caption{Impact of response function choice in $\mathcal{G}_2$ on rationality parameters in $\mathcal{G}_1$. Maxmax (MX) as the solution concept in $\mathcal{G}_2$ lead to higher precision parameter estimates.}
\label{tab:rq3}
\end{table}
In general, similar to the results in Fig. \ref{fig:delta_u_distr}, QlkR models show higher values of the precision parameter as well, thus reflecting better performance as a model of behaviour in level-1 games, i.e. for selection of manoeuvres.  QlkR:MX\textsubscript{S(1+B)} (QlkR model with maxmax response in $\mathcal{G}_2$ with bounds sampling) show highest value of the precision parameter, $\lambda=22.1\pm 0.22$ (Fig. \ref{fig:lambda_chart}a). Next, we evaluate model fit using Akaike information criterion (AIC) values, which are noted in Fig. \ref{fig:lambda_chart}(a) in brackets. Since AIC is an evaluation of model fit rather than predictive accuracy, the log likelihood values used in AIC calculation was performed over the entire dataset instead of just the training set. QlkR model with bounds sampling of trajectories have lowest AIC values (-225.3 and -195.6  for QlkR:MM\textsubscript{S(1+B)} and QlkR:MX\textsubscript{S(1+B)} respectively), indicating the best fit among the models based on this criterion.\par
Alternatively, model selection can also be guided by their predictive power in unseen situations. For evaluation based on this criterion, we use random subsampling with 75:25 training and testing split and 30 runs. The model parameters are estimated based on the observations in the training set, and the predictive accuracy is measured in two ways. First, the predictive accuracy of the models is measured when the solution is in mixed strategies; which is evaluated on the basis of the log likelihood of the observed actions in the testing set. Second, the models are also evaluated based on their accuracy of pure strategy solutions. 
Fig. \ref{fig:lambda_chart}(b) shows the boxplot of sum log likelihood of the observed $\mathcal{G}_{1}$ actions in the testing set as predicted by each model across 30 runs. The sum log likelihood is calculated based on the likelihood of the observed actions in the test set as predicted by the Quantal Response model with the estimated precision parameter ($\hat{\lambda}$). The set of available actions for the games are often different (since the available actions depend on the state of the road user), and therefore to standardize the analysis process, the likelihood was calculated over the domain of utilities based on a continuous negative exponential distribution rather than over the actions as is often done in estimation of parameter of Quantal Best Response. This process of estimating the likelihood is invariant to different sets of actions, but still keeps the main model of Quantal Best Response intact. However, note that the likelihood values in this transformed model can be greater than 1. More specifically, the likelihood value of an individual observed action was calculated using the formula $\pi^{\text{QBR}(a_{-i}^{*},\hat{\lambda_{i}})}(a_{i}^{o}) = \hat{\lambda}e^{-\hat{\lambda}\Delta u}$ where $\Delta u= u_{i}(a_{i}^{*},a_{-i}^{*}) - u_{i}(a_{i}^{o},a_{-i}^{*})$ is the difference in utility between the game's solution and the selected action, $\hat{\lambda}$ is the estimate of the precision parameter based on the generalized linear model, and the sum log likelihood was calculated with the formula $\sum\log(\pi^{\text{QBR}(a_{-i}^{*},\hat{\lambda_{i}})}(a_{i}^{o}))$. \par
\begin{figure*}[t]%
\centering
\includegraphics[width=0.8\textwidth]{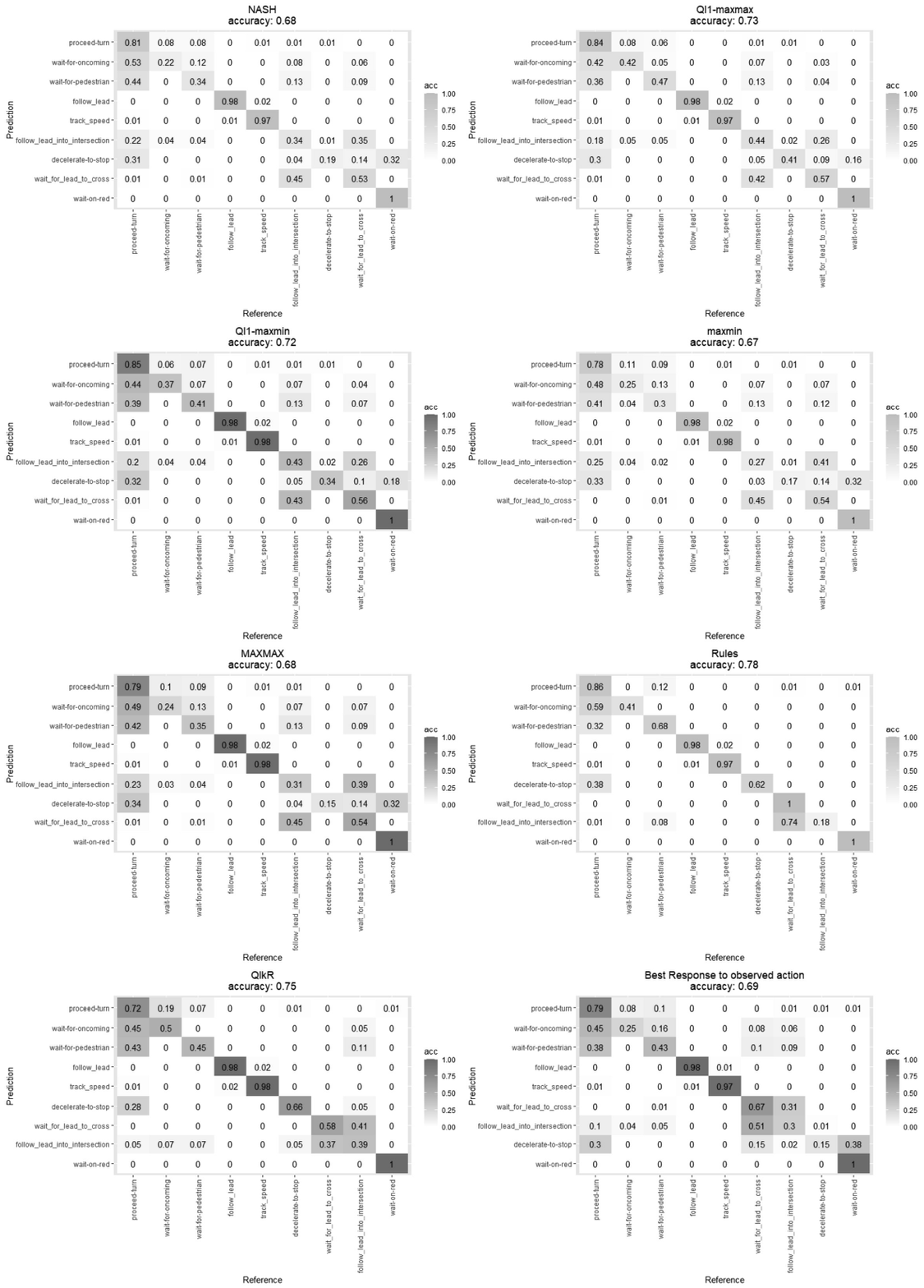}%
\caption{Confusion matrix of the pure strategy solutions of level-1 games with S(1) sampling of models with respect to the ground truth maneuver. }
\label{fig:ch-oneshot-l1_accu}
\end{figure*}

Next, we further compare the models with respect to their pure strategy solutions. Although the mixed strategy solutions of the models (as expressed through the precision parameter) give a good understanding of how well the models capture naturalistic behaviour, when it comes to using the behaviour models as a behaviour planner in an AV, it is important to evaluate them also with respect to the pure strategy solutions, since an AV can only execute a single action at a time rather than a mixed one. Fig. \ref{fig:ch-oneshot-l1_accu} shows the multi-class confusion matrix of the predicted manoeuvres ($\mathcal{G}_1$ solutions) of each model. Instead of all 30 models, we select a cross sectional sample of the five models that are based on S(1) sampling of trajectories. In addition to the models already presented in the  paper, we also include a model solely for the sake of comparison. In this model, labelled `best response to observed action', as the name suggests, agents simply best respond to the observed manoeuvre of other agents in the game that was played in the previous time step. Based on the data, we see that the accuracy varies significantly across different manoeuvres. The mean accuracy is highest for the rule-based model (78\%) followed by QlkR model (75\%), which is not surprising since the dataset contains many situations that do not involve strategic reasoning which the strategic behaviour models are good for, for example, approaching the intersection before deciding whether to take the turn. With respect to specific manoeuvres, the behaviour models fare better for the decision of whether to tail a lead vehicle into an intersection or not (follow-lead-into-intersection). On the other hand, the rule based model does better with respect to waiting for oncoming vehicles and pedestrians. In the next  paper, we will revisit this comparison again by focusing on situations that involve higher chance of strategic reasoning as well as some techniques that improve the overall accuracy of the models.\par
Overall, these results indicate that based on the three evaluation criteria combined (precision parameter, AIC, and predictive performance) and all the models studied, QlkR model where players best respond to the belief that others will follow the traffic rules, especially with bounds sampling of trajectories, is the better model of decision making at the level of manoeuvres for both pure strategy (comparison based on best response) and mixed strategy responses (comparison based on Quantal response).\par

\begin{sidewaystable}[ht]
\centering
\resizebox{0.6\width}{!}{%
\begin{tabular}{@{}p{2.5cm}|p{1.76cm}|p{1.76cm}|p{1.76cm}|p{1.76cm}|p{1.76cm}|p{1.76cm}|p{1.76cm}|p{1.76cm}|p{1.76cm}|p{1.76cm}|p{1.76cm}|p{1.76cm}|p{1.76cm}|p{1.76cm}|p{1.76cm}@{}}%
  \toprule
STATE FACTOR & Ql0: BR\_S(1) & PNE-QE: BR\_ S(1+B) & PNE-QE: BR\_ S(1+G) & PNE-QE: MM\_ S(1+B) & PNE-QE: MM\_ S(1+G) & PNE-QE \_S(1) & Ql0: BR: BR\_S(1+B) & Ql0: BR: BR\_S(1+G) & Ql0: BR: MM\_S(1+B) & Ql0: BR: MM\_S(1+G) & Ql0: MM: BR\_S(1+B) & Ql0: MM: BR\_S(1+G) & Ql0: MM: MM\_S(1+B) & Ql0: MM: MM\_S(1+G) & Ql0: MM\_ S(1) \\ 
  \midrule
  SEGMENT&&&&&&&&&&&&&&&\\%
exec-left-turn & 9.61 & 11.31 & 10.03 & 7.05 & 7.60 & 9.54 & 13.22 & 11.31 & 7.56 & 7.82 & 12.37 & 10.77 & 7.32 & 7.62 & 9.16 \\ 
exec-right-turn & 10.57 & 9.19 & 9.31 & 6.73 & 7.38 & 10.64 & 32.04 & 28.85 & 16.34 & 15.88 & 10.26 & 9.87 & 7.08 & 7.05 & 10.53 \\ 
OTHER    LANES & 9.07 & 12.20 & 10.03 & 7.50 & 7.31 & 9.02 & 14.12 & 11.56 & 7.72 & 7.86 & 12.72 & 10.94 & 7.55 & 7.71 & 9.04 \\ 
prep-left-turn & 5.29 & 10.33 & 8.84 & 4.25 & 4.76 & 5.43 & 10.93 & 8.88 & 4.11 & 4.45 & 10.57 & 8.61 & 4.07 & 4.41 & 5.28 \\ 
prep-right-turn & 9.69 & 9.82 & 8.37 & 5.29 & 5.44 & 9.20 & 21.37 & 15.04 & 6.73 & 6.99 & 10.79 & 8.64 & 5.40 & 5.47 & 9.17 \\ 
\midrule%
  NEXT CHANGE&&&&&&&&&&&&&&&\\%
  $<$10-G & 26.68 & 28.29 & 20.29 & 18.77 & 15.31 & 25.25 & 30.78 & 24.15 & 19.09 & 17.75 & 29.73 & 23.53 & 18.95 & 17.50 & 24.46 \\ 
  $<$10-Y/R & 5.81 & 10.86 & 9.70 & 5.10 & 5.46 & 5.87 & 14.41 & 11.12 & 5.29 & 5.44 & 11.61 & 9.71 & 5.02 & 5.22 & 5.75 \\ 
  $>$EQ    10-G & 16.04 & 21.68 & 16.54 & 16.61 & 14.61 & 16.17 & 23.07 & 19.42 & 17.22 & 16.46 & 22.11 & 18.98 & 16.85 & 16.07 & 15.97 \\ 
  $>$EQ    10-Y/R & 5.61 & 8.03 & 7.00 & 4.29 & 4.64 & 5.56 & 9.99 & 8.08 & 4.52 & 4.81 & 8.52 & 7.22 & 4.31 & 4.57 & 5.56 \\ 
  \midrule
   SPEED&&&&&&&&&&&&&&&\\%
  HIGH & 3.58 & 6.79 & 6.74 & 6.85 & 6.77 & 3.58 & 6.79 & 6.74 & 6.85 & 6.79 & 6.79 & 6.75 & 6.85 & 6.79 & 3.58 \\ 
  LOW & 8.06 & 11.47 & 9.57 & 6.48 & 6.59 & 8.05 & 13.64 & 11.17 & 6.76 & 7.01 & 12.05 & 10.26 & 6.50 & 6.73 & 8.02 \\ 
  MEDIUM & 23.08 & 13.83 & 13.35 & 9.81 & 10.90 & 21.96 & 20.22 & 16.57 & 11.62 & 12.16 & 14.80 & 13.40 & 10.05 & 10.73 & 19.55 \\ \midrule
  PEDESTRIAN&&&&&&&&&&&&&&&\\%
  N & 8.30 & 12.42 & 10.43 & 6.89 & 7.10 & 8.35 & 14.18 & 11.90 & 7.18 & 7.52 & 12.98 & 11.16 & 6.92 & 7.24 & 8.33 \\ 
  Y & 8.38 & 10.66 & 8.93 & 6.25 & 6.32 & 8.26 & 13.52 & 10.76 & 6.62 & 6.79 & 11.26 & 9.53 & 6.29 & 6.46 & 8.18 \\ 
  \midrule
  NO. RELEV. VEHICLE&&&&&&&&&&&&&&&\\%
  $\leqslant2$ & 9.56 & 8.80 & 8.70 & 6.74 & 7.09 & 8.97 & 11.91 & 10.19 & 7.95 & 7.59 & 10.46 & 9.34 & 7.53 & 7.23 & 9.45 \\ 
  $>2$& 4.58 & 20.21 & 21.07 & 6.00 & 3.40 & 6.47 & 32.81 & 24.70 & 3.40 & 3.23 & 29.75 & 22.15 & 3.37 & 3.19 & 4.54 \\ 
  $=0$ & 19.74 & 16.77 & 13.94 & 7.60 & 8.12 & 18.98 & \underline{\textbf{22.96}} & 20.56 & 15.44 & 16.03 & 18.99 & 17.84 & 14.00 & 14.61 & 19.56 \\ 
   \bottomrule
\end{tabular}}
\caption{Mean precision parameter ($\lambda_{i,b}$) of the behaviour models for each state variable across all games.}
\label{tab:x_l_table1}
\end{sidewaystable}

\begin{sidewaystable}[ht]
\centering
\resizebox{0.6\width}{!}{%
\begin{tabular}{@{}p{2.5cm}|p{1.76cm}|p{1.76cm}|p{1.76cm}|p{1.76cm}|p{1.76cm}|p{1.76cm}|p{1.76cm}|p{1.76cm}|p{1.76cm}|p{1.76cm}|p{1.76cm}|p{1.76cm}|p{1.76cm}|p{1.76cm}|p{1.76cm}@{}}
  \toprule
 STATE FACTOR& Ql1: BR: BR\_ S(1+B) & Ql1: BR: BR\_ S(1+G) & Ql1: BR: MM\_ S(1+B) & Ql1: BR: MM\_ S(1+G) & Ql1: BR\_ S(1) & Ql1: MM: BR\_ S(1+B) & Ql1: MM: BR\_ S(1+G) & Ql1: MM: MM\_ S(1+B) & Ql1: MM: MM\_ S(1+G) & Ql1: MM\_ S(1) & QlkR: BR-R: BR\_ S(1+B) & QlkR: BR-R: BR\_ S(1+G) & QlkR: BR-R: MM\_S(1+B) & QlkR: BR-R: MM\_ S(1+G) & QlkR: BR\_S(1) \\ 
  \midrule
  SEGMENT&&&&&&&&&&&&&&\\%
exec-left-turn & 12.66 & 10.93 & 7.39 & 7.67 & 9.56 & 12.66 & 10.93 & 7.39 & 7.67 & 9.56 & 15.43 & 13.43 & 11.70 & 10.49 & \underline{17.83} \\ 
exec-right-turn & 15.47 & 13.97 & 9.00 & 8.84 & 10.67 & 15.47 & 13.97 & 9.00 & 8.84 & 10.68 & \underline{\textbf{22.67}} & 21.20 & 13.40 & 12.69 & 21.67 \\ 
OTHER    LANES & 13.30 & 11.21 & 7.61 & 7.77 & 9.10 & 13.30 & 11.21 & 7.61 & 7.77 & 9.09 & 31.16 & \underline{\textbf{29.36}} & 24.03 & 22.45 & 22.44 \\ 
prep-left-turn & 10.76 & 8.72 & 4.08 & 4.42 & 5.27 & 10.76 & 8.72 & 4.08 & 4.42 & 5.27 &\underline{\textbf{ 20.06}} & 19.13 & 14.48 & 14.17 & 14.84 \\ 
prep-right-turn & 14.07 & 10.64 & 5.72 & 5.87 & 9.41 & 14.07 & 10.64 & 5.72 & 5.87 & 9.41 & 23.85 & 22.60 & 13.53 & 13.03 & \underline{\textbf{25.08}} \\ 
\midrule%
  NEXT CHANGE&&&&&&&&&&&&&&\\%
$<$10-G & 30.06 & 23.96 & 19.02 & 17.57 & 26.19 & 30.06 & 23.96 & 19.02 & 17.57 & 26.17 & 103.14 & 87.35 & 81.60 & 67.60 & \underline{\textbf{93.19}} \\ 
$<$10-Y/R & 12.77 & 10.24 & 5.08 & 5.29 & 5.70 & 12.77 & 10.24 & 5.08 & 5.29 & 5.70 & \underline{\textbf{24.59}} & 21.73 & 13.75 & 12.86 & 12.24 \\ 
$>$EQ    10-G & 22.51 & 19.18 & 17.09 & 16.23 & 16.27 & 22.51 & 19.18 & 17.09 & 16.23 & 16.21 & \underline{\textbf{76.81}} & 69.05 & 73.47 & 62.97 & 62.67 \\ 
$>$EQ    10-Y/R & 9.13 & 7.57 & 4.38 & 4.65 & 5.60 & 9.13 & 7.57 & 4.38 & 4.65 & 5.60 & 15.30 & \underline{\textbf{14.52} }& 11.28 & 10.78 & 12.22 \\ 
\midrule
   SPEED&&&&&&&&&&&&&&\\%
HIGH& 6.79 & 6.74 & 6.85 & 6.77 & 3.58 & 6.79 & 6.74 & 6.85 & 6.77 & 3.58 & 9.31 & \underline{\textbf{9.43}} & 6.03 & 7.35 & 4.34 \\ 
LOW& 12.72 & 10.64 & 6.58 & 6.83 & 8.07 & 12.72 & 10.64 & 6.58 & 6.83 & 8.07 & \underline{\textbf{25.42}} & 23.70 & 18.61 & 17.50 & 20.42 \\ 
MEDIUM& 16.77 & 14.63 & 10.59 & 11.20 & 22.28 & 16.77 & 14.63 & 10.59 & 11.20 & 22.28 & 28.99 & 24.97 & 26.33 & 22.19 & \underline{\textbf{31.33}} \\
\midrule
  PEDESTRIAN&&&&&&&&&&&&&&\\%
N& 13.50 & 11.46 & 7.01 & 7.33 & 8.34 & 13.50 & 11.46 & 7.01 & 7.33 & 8.34 & \underline{\textbf{27.87}} & 25.47 & 19.67 & 18.44 & 20.63 \\ 
Y& 12.18 & 10.04 & 6.38 & 6.57 & 8.33 & 12.18 & 10.04 & 6.38 & 6.57 & 8.33 & 23.02 & \underline{\textbf{21.77}} & 17.96 & 16.79 & 20.59 \\ 
\midrule
  NO. RELEV. VEHICLE&&&&&&&&&&&&&&\\%
$\leqslant2$& 11.10 & 9.70 & 7.67 & 7.36 & 9.59 & 11.10 & 9.70 & 7.67 & 7.36 & 9.59 & \underline{\textbf{23.27}} & 21.03 & 18.88 & 17.33 & 20.16 \\ 
$>2$& 30.80 & 23.14 & 3.38 & 3.20 & 4.55 & 30.80 & 23.14 & 3.38 & 3.20 & 4.55 & 48.66 & \underline{\textbf{56.83}} & 18.90 & 19.07 & 15.62 \\ 
$=0$& 20.16 & 18.63 & 14.19 & 14.94 & 19.30 & 20.16 & 18.63 & 14.19 & 14.94 & 19.13 & 22.67 & 21.61 & 19.17 & 18.21 & \underline{\textbf{80.47}} \\ 
   \bottomrule
\end{tabular}}
\caption{Mean precision parameter ($\lambda_{i,b}$) of the behaviour models for each state variable across all games (continued).}
\label{tab:x_l_table2}
\end{sidewaystable}

\emph{RQ2. How do state factors influence the precision parameters in the games?} In this research question, we study the impact of the state factors on the precision parameter. The state factors are shown in the first column of Tables \ref{tab:x_l_table1} and \ref{tab:x_l_table2}. Most state factors are self explanatory; NEXT\_CHANGE refers to the next change in the traffic signal and time in seconds till the change occurs, RELEV\_VEHICLE refers to the type of relevant vehicle in the game, for example, whether there is a lead vehicle present or other vehicles in conflict which are not lead vehicles. The table shows the mean precision parameter of the behaviour models for each state factor variable. Since $\lambda_{i,b}$ depends on the state $X_{i}$, which is a vector of the five categorical state factors, each row in the table shows the mean precision parameter for situations with the corresponding state factor value, but in isolation; i.e. without taking into account the interaction between the state factors like in the predictive \emph{glm} model. For each state factor value, the value corresponding to the highest precision parameter is underlined. As expected from the previous results, for most state factor variables, QlkR models have the highest precision parameter values. When we compare the values of the precision parameter in Tables \ref{tab:x_l_table1} and \ref{tab:x_l_table2} with the values of Fig. \ref{fig:lambda_chart} (b), we observe that there is much more variation within individual models depending on the agent's state compared to the variation between different models. \par

\emph{RQ3. How does the choice of the response function in the lower action level game $\mathcal{G}_2$ affect the higher level solutions in $\mathcal{G}_1$?} As part of this research question, we analyse the impact on the estimate of the precision parameter based on the choice of the solution concept in the lower action level game $\mathcal{G}_2$. For the six possible combinations of the metamodel and the solution concept in $\mathcal{G}_{1}$, namely, Ql0:MX, Ql0:MM, Ql1:MX, Ql1:MM, PNE-QE, and QlkR, Table \ref{tab:rq3} shows the relative change in the precision parameter of action level games $\mathcal{G}_1$ based on the choice of the response function in $\mathcal{G}_2$. All estimates were found to be significant at $p=0.05$ based on Dunn's pairwise comparison test after Kruskal-Wallis test indicated significant within group difference. We see that choosing maxmax as the solution concept in $\mathcal{G}_2$ consistently results in a better precision parameter after controlling for the model and the solution concept in $\mathcal{G}_1$.

\section{Conclusion}
We formalise the concept of a hierarchical game and develop various solution concepts that can be applied to a hierarchical game by adapting popular behavioural game theoretic metamodels (Qlk and PNE-QE). In the context where games are constructed to model naturalistic scenarios, modellers are faced with multiple choices, and this  paper shows different ways in which strategic and non-strategic models can be applied to solve a hierarchical game.  We evaluated the behaviour models based on a large dataset of human driving at a busy urban intersection. The results show that among the behaviour models evaluated, modelling driving behaviour as a model where drivers best respond to other drivers with the belief that everyone else will follow the rules is the superior model of manoeuvre selection. As a design choice, constructing the action space of the games with bounds sampling of trajectories provides the best fit to naturalistic driving behaviour. However, if computational efficiency is a concern, then modellers do not lose much performance if they use a single prototype trajectory as a method of constructing actions in a hierarchical game. Furthermore, choosing maxmax as a solution concept for solving the game of trajectories results in higher precision parameter values compared to a maxmin model. The work undertaken in this  paper provides practical insight for practitioners interested in modelling interactive human decision making in traffic for autonomous vehicles.

\FloatBarrier
\begin{quote}
\begin{small}
\bibliography{main}
\end{small}
\end{quote}

\end{document}